\documentclass[opre,nonblindrev]{informs3} 

\OneAndAHalfSpacedXI 

\RequirePackage[numbers]{natbib}
\RequirePackage[colorlinks,citecolor=blue,linkcolor=blue,urlcolor=blue]{hyperref}
\RequirePackage{graphicx}
\usepackage{float}
\usepackage{caption}
\usepackage{subcaption}
\usepackage{comment}
\usepackage{enumitem}
\usepackage{algorithm}
\usepackage{algpseudocode}
\usepackage{tikz}

\usepackage{natbib}
 \bibpunct[, ]{(}{)}{,}{a}{}{,}%
 \def\bibfont{\small}%
\TheoremsNumberedThrough     
\ECRepeatTheorems

\EquationsNumberedThrough    

\newcommand\prs[1]{\left(#1\right)}
\newcommand\sbk[1]{\left[#1\right]}
\newcommand{\E}{\mathbb{E}}

\newcommand{\gti}{\rightarrow\infty}

\newcommand{\var}{\mathrm{Var}}
\newcommand{\cov}{\mathrm{Cov}}

\begin{document}


\RUNAUTHOR{
Qu, Namkoong, and Zeevi
}

\RUNTITLE{A Broader View of Thompson Sampling}

\TITLE{A Broader View of Thompson Sampling}

\ARTICLEAUTHORS{%
\AUTHOR{Yanlin Qu}
\AFF{Columbia Business School, \EMAIL{qu.yanlin@columbia.edu}} 
\AUTHOR{Hongseok Namkoong}
\AFF{Columbia Business School, \EMAIL{namkoong@gsb.columbia.edu}}
\AUTHOR{Assaf Zeevi}
\AFF{Columbia Business School, \EMAIL{assaf@gsb.columbia.edu}}
} 

\ABSTRACT{%
  Thompson Sampling is one of the most widely used and studied bandit algorithms, known for its simple structure, low regret performance, and solid theoretical guarantees. Yet, in stark contrast to most other families of bandit algorithms, the exact mechanism through which  posterior sampling (as introduced by Thompson) is able to ``properly'' balance exploration and exploitation, remains a mystery. In this paper, we show that the core insight to address this question stems from recasting Thompson Sampling as an online optimization algorithm.
  To distill this, we introduce a suitable time invariant notion of regret that leads to a stationarized bandit problem, and a stationary Bellman-optimal policy. We then show  that Thompson Sampling admits an online optimization form that mimics the structure of the aforementioned  Bellman-optimal policy, where ``greediness" is regularized by a measure of residual uncertainty. This new lens of online optimization allows both a better understanding of  Thompson Sampling dynamics, as well as  a principled manner for policy improvement that  mimics the Bellman-optimal benchmark.
}

%



\KEYWORDS{Multi-armed Bandit, Regret Minimization, Bellman Equation, Thompson Sampling, Online Optimization, Policy Improvement} 
\maketitle

\section{Introduction}
\textbf{Background and motivation.} Thompson Sampling  is a heuristic Bayesian algorithm, introduced by \cite{thompson1933likelihood} in the context of solving treatment allocation in medical trials; the objective is to maximize patient outcomes while simultaneously learning the best treatment. (This motivating application has since been abstracted to what we recognize today as the multi-armed bandit (MAB) problem.) Thompson Sampling  is initialized with a prior distribution (belief) over system parameters, and then proceeds in each round to sample from a posterior distribution, the updated belief over problem parameters based on data observed to that point, selecting the treatment (arm) that is perceived to be optimal in the sampled environment. In other words, its basic design interleaves a simple  sampling based approach with Bayes rule. 

While Thompson Sampling remained obscure throughout the 20th century, the MAB problem has attracted significant attention ever since its first definitive formulation by \cite{robbins1952some}. In addition to formalizing the problem, that paper made a foundational observation about the basic tension between {\it exploration} and {\it exploitation} that is central to online learning: any procedure aiming to maximize long-run average reward must explore all arms infinitely often to avoid ``missing" the optimal arm.  
In continuation of this principle, a landmark paper by \cite{lai1985asymptotically} introduced the notion of {\it regret}, the loss incurred by any sequential decision rule relative to an oracle that knows the identity of the best arm, and proposed a policy that carefully assigns (infinitely many) pulls to each arm to achieve the minimal possible growth rate of regret. This policy was later simplified to the ubiquitous Upper Confidence Bound (UCB) algorithm, popularized by \cite{auer2002finite}. 

Close to a decade after the \cite{auer2002finite} paper, Thompson Sampling was finally resurrected, triggered by several studies that indicated  remarkably strong empirical performance \citep[e.g.,][]{scott2010modern,chapelle2011empirical}, often rivaling or even surpassing that of UCB. Since then, practitioners have applied Thompson Sampling across a wide range of domains, including online advertising \citep[e.g.,][]{agarwal2013computational}, recommendation systems \citep[e.g.,][]{kawale2015efficient}, and website optimization \citep[e.g.,][]{hill2017efficient}. Concurrently, a substantial body of theoretical work has developed, focusing on bounding the regret of Thompson Sampling, and essentially showing that it achieves the goal of long-term regret minimization; see the  frequentist regret bounds in \cite{agrawal2012analysis, agrawal2013further} and Bayesian regret bounds in \cite{russo2014learningPS, russo2016information}. 

While the aforementioned theory introduced several innovative ideas and technical tools that extend beyond Thompson Sampling, it falls short of elucidating the key optimization principle or at least the explicit exploration-exploitation tradeoffs that guide Thompson Sampling. This stands in stark contrast to upper confidence bound policies (in particular the simplified version in \cite{auer2002finite}), and variants thereof such as explore-then-commit, epsilon-greedy and the like (see, e.g., \cite{lattimore2020bandit}), where  exploration and exploitation considerations are quite transparent and in fact guide the design of these algorithms. 
  
To that end, it is worth noting that neither Thompson Sampling nor the UCB family are derived from standard optimization  principles such as dynamic programming \citep{bellman1957dp}. A key illustration of the latter is the Gittins index policy \citep{gittins1979bandit}, which formulates the Bayesian version of the MAB problem as a Markov decision process (MDP), and derives the optimal policy that maximizes expected cumulative discounted reward. While discounting simplifies the problem by making it stationary, in contrast to the widely studied traditional finite horizon regret setting,  it also results in a significant deviation from the intuitive principle laid out by \cite{robbins1952some}. Specifically, the Gittins index policy may pull the optimal arm only finitely many times, and hence fail to ``identify'' it, resulting in performance dramatically inferior to that of Thompson Sampling (and UCB) absent discounting.  

In this paper, akin to Gittins, we aim to harness Bellman's more principled approach to shed further light on the optimization considerations underlying Thompson Sampling. But toward that end, and to remain within the traditional finite horizon regret formulation, where the success of Thompson Sampling was established and validated, we depart from Gittins' infinite horizon discounted reward formulation. In lieu of that, we propose a different form of stationarization, which is more ``faithful"  to Robbins' original principle, and show that through this lens, Thompson Sampling takes the form of an online optimization algorithm that at each step balances between greediness and a measure of residual uncertainty which serves as a {\it regularizer}. Beyond addressing the core question of what Thompson Sampling optimizes, it also provides a principled framework for further understanding and improvement to this important class of posterior sampling algorithms.

\textbf{Main contributions and overview of key ideas.} 
We first describe our proposed notion of ``faithful"  stationarization of the long-term regret minimization problem, as this holds the key to explaining Thompson Sampling through an optimization lens. 
For simplicity of exposition, and to stay true to Thompson's original 1933 setup, we consider a two-armed bandit with independent arms, each generating random rewards when pulled. (The $K$-armed case is discussed at the end of Section \ref{sec_pi}.) The learner's goal is maximizing cumulative reward, or equivalently minimizing cumulative regret
\begin{equation}
\label{eqn_regret}
\mathcal{R}_T(Q;\pi_0)=\E_{\pi_0}\sbk{\sum_{t=0}^{T-1}\sbk{\mathrm{max}(\theta_1,\theta_2)-\theta_{A_t}}},
\end{equation}
where $Q$ is a policy, $T$ is a finite time horizon, $\theta_k$ is the mean reward of arm $k$, and $A_t$ is the arm chosen at time $t$. In the Bayesian setting, the expectation is taken over the randomness of interaction (rewards observed and arms pulled) and the environment, as the unknown parameter $\theta=(\theta_1,\theta_2)$ is drawn from a prior distribution $\pi_0$ before the game begins. 

As noted by \citep{gittins1979bandit}, this bandit problem becomes a Markov decision process (MDP) when posterior distributions $\pi_1,\pi_2,...$ are viewed as states. For MDPs, perhaps the most principled framework for optimizing performance is dynamic programming, typically expressed through Bellman equations. In particular, stationary Bellman equations (e.g., infinite horizon with discounting) are typically more tractable than their non-stationary counterparts (e.g., finite horizon).
For example, maximizing cumulative discounted reward
\begin{equation}
\label{eqn_discount}
\E_{\pi_0}\sbk{\sum_{t=0}^\infty \gamma^t\theta_{A_t}},\;\;\gamma\in(0,1)
\end{equation}
leads to the elegant optimal policy known as the Gittins index. However, as mentioned earlier, discounted \eqref{eqn_discount} and non-discounted \eqref{eqn_regret} are fundamentally different objectives. This discrepancy has significant consequences. In fact, the Gittins index policy, despite maximizing \eqref{eqn_discount}, can suffer linear regret, i.e., \eqref{eqn_regret} grows linearly in $T$; see, e.g., \cite{rothschild1974two}. 

To obtain a stationary Bellman equation that is faithful to minimizing \eqref{eqn_regret}, we consider minimizing cumulative {\it squared regret} 
\[
\mathcal{R}^2(Q;\pi_0)=\E_{\pi_0}\sbk{\sum_{t=0}^{\infty}r^2(q_t;\pi_t)},
\]
where $q_t$ is the distribution of $A_t|\pi_t$ under policy $Q$, and $r(q_t;\pi_t)=\E_{\pi_t}\sbk{\mathrm{max}(\theta_1,\theta_2)-\theta_{A_t}}$ is the expected next-round regret (given $\pi_t$). This new objective is aligned with the original one in the sense that minimizing $\mathcal{R}^2(Q;\pi_0)$ minimizes the following regret bound (proved later in the paper)
\begin{equation}
\label{eqn_bound}
\mathcal{R}_T(Q;\pi_0)\leq\sqrt{\mathcal{R}^2(Q;\pi_0)\cdot T}.
\end{equation}
The corresponding stationary Bellman equation is
\[
V(\pi_t)=\min_{q_t}\sbk{r^2(q_t;\pi_t)+q_t\cdot\E_{\pi_t}\sbk{V(\pi_{t+1})|A_t=\cdot}},
\]
where $V(\pi_t)=\mathcal{R}^2(Q^\mathrm{R2};\pi_t)$ is the minimal squared regret achieved by the $\mathcal{R}^2$-optimal policy $Q^\mathrm{R2}$.
After a suitable change of variables, $Q^\mathrm{R2}$ admits an intuitive online optimization structure 
\[
x_t^\mathrm{R2}=\underset{x_t}{\mathrm{argmin}}\sbk{\prs{\E_{\pi_t}\max(\theta_1,\theta_2)-x_t}^2+\nu^\mathrm{R2}(\pi_t)x_t},
\]
where the expected next-round reward $x_t=q_t\cdot\E_{\pi_t}\theta$ is chosen to minimize the instantaneous squared regret subject to linear regularization. The regularizer $\nu^\mathrm{R2}(\pi_t)$ (defined later in the paper) will be seen to measure the {\it current tension} (at each time $t=1,2,\ldots$) between exploration and exploitation.

Remarkably, despite being introduced as a heuristic without an explicit optimization objective, Thompson Sampling can also be expressed in the same online optimization form as above, namely 
\[
x_t^\mathrm{TS}=\underset{x_t}{\mathrm{argmin}}\sbk{\prs{\E_{\pi_t}\max(\theta_1,\theta_2)-x_t}^2+\nu^\mathrm{TS}(\pi_t)x_t},
\]
where the regularizer now becomes $\nu^\mathrm{TS}(\pi_t)=\cov_{\pi_t}(\theta_1-\theta_2,\mathrm{sign}(\theta_1-\theta_2))$. This is the familiar notion of ``biserial'' covariance \citep{pearson1909new} which measures the {\it remaining uncertainty} about the ``better" arm (on the scale of regret), thereby quantifying the exploration logic underlying Thompson Sampling as an instance of uncertainty-driven regularization. Recall that the Bellman equation suggests a tension-driven regularization.
Motivated by this difference, we compare Thompson Sampling against the Bellman-optimal benchmark to understand and improve the heuristic algorithm in a principled manner:

\begin{figure}[ht]
    \centering
    \begin{minipage}{0.49\textwidth}
        \centering
        \includegraphics[width=0.9\linewidth]{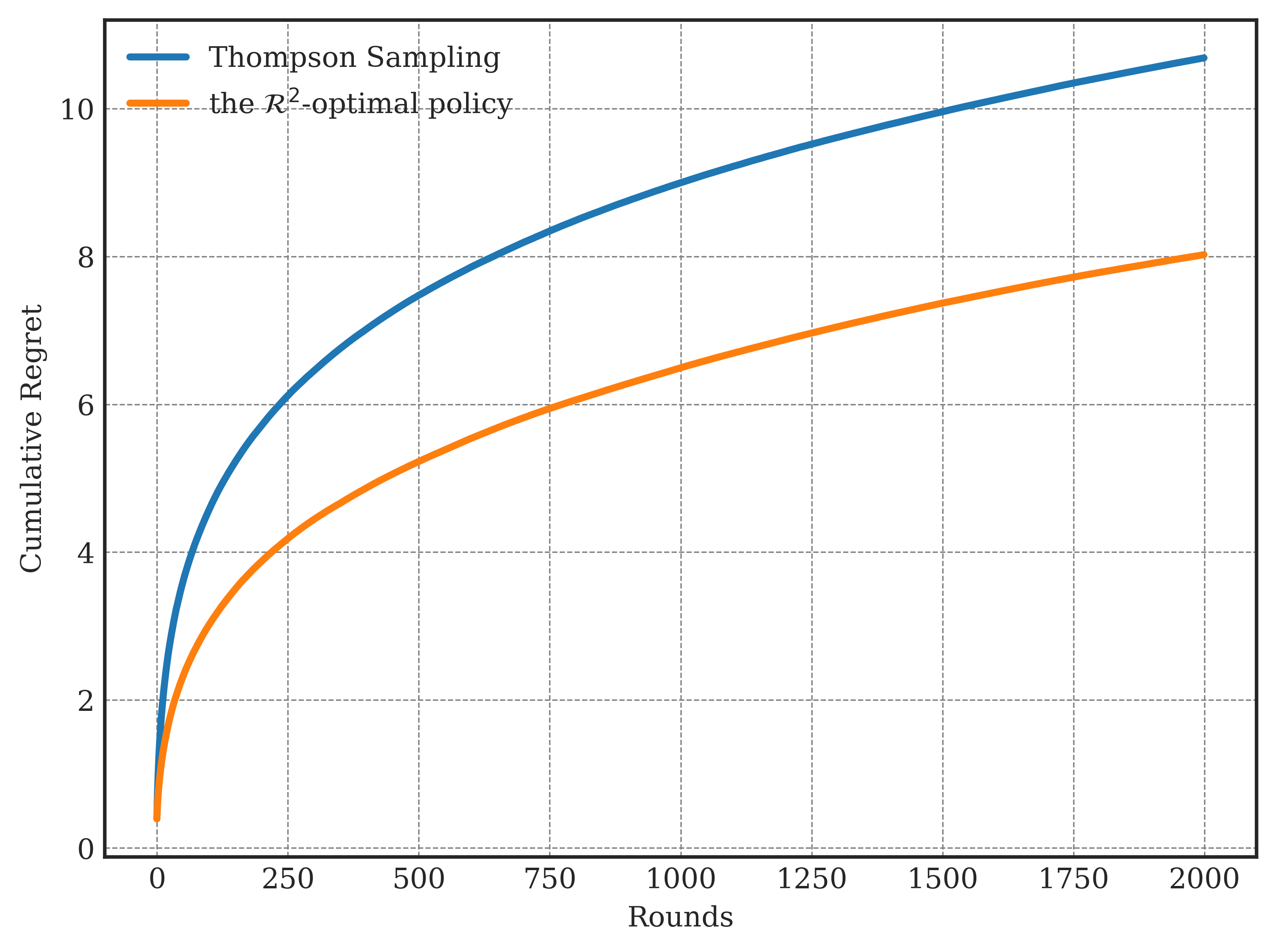}
    \end{minipage}
    \hfill
    \begin{minipage}{0.49\textwidth}
        \centering
        \includegraphics[width=0.9\linewidth]{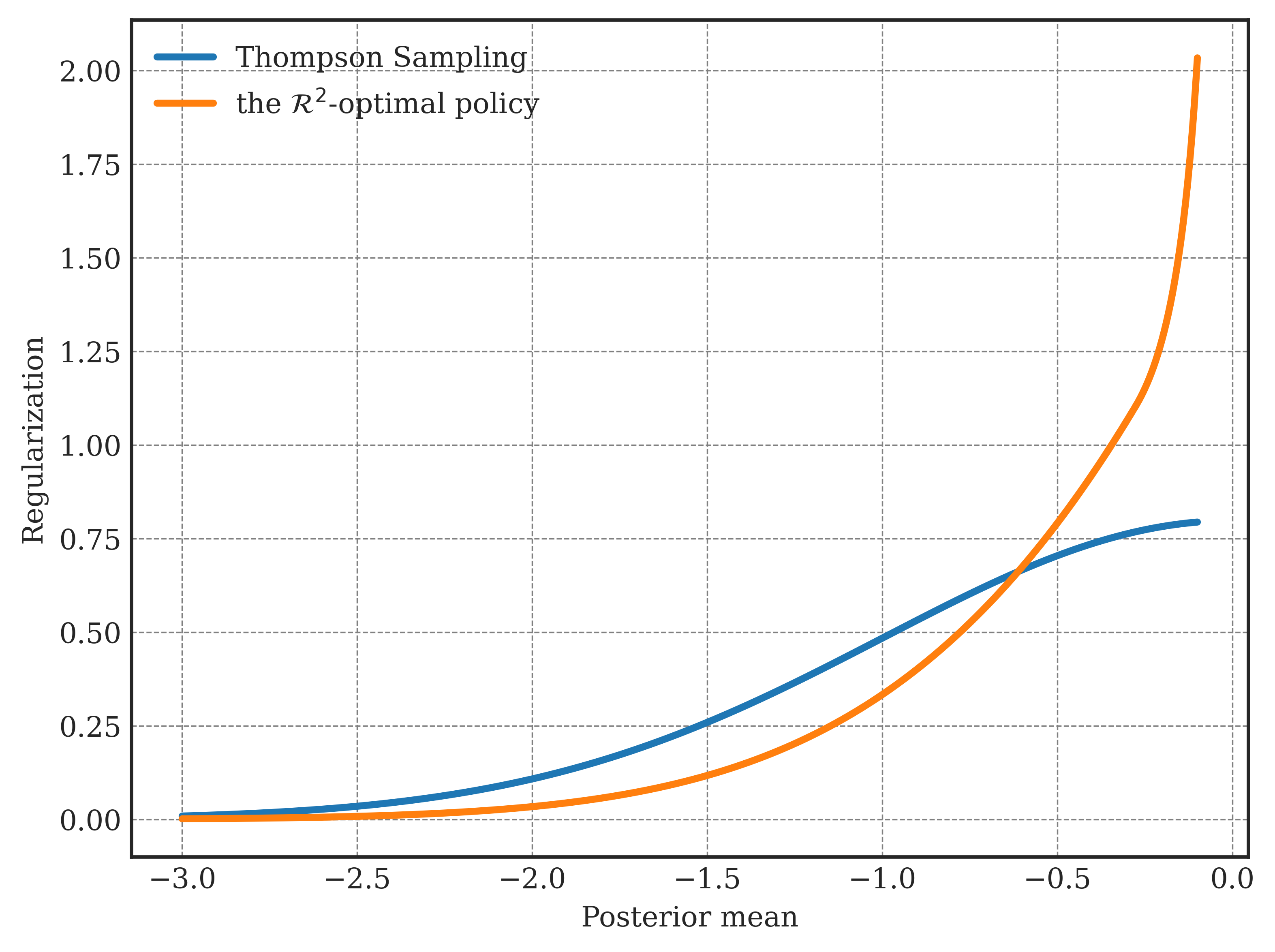}
    \end{minipage}

    \caption{Thompson Sampling and the $\mathcal{R}^2$-optimal policy play a Gaussian bandit with reward variance 1. Left: Comparing their cumulative regret $\mathcal{R}_T(Q^\text{TS};\pi_0)$ vs. $\mathcal{R}_T(Q^\text{R2};\pi_0)$ where $\pi_0=N(0,1)\times\delta_0$ (20K trials). Right: Comparing their regularizers $\nu^\text{TS}(N(\mu,1)\times\delta_0)$ vs. $\nu^\text{R2}(N(\mu,1)\times\delta_0)$ where $\mu\in[-3,0]$.
    }
    \label{figure_intro}
\end{figure}

\begin{itemize}
    \item The $\mathcal{R}^2$ Bellman-optimal policy $Q^\mathrm{R2}$ achieves the best possible constant in the regret bound \eqref{eqn_bound}, and thus enjoys a stronger theoretical guarantee than Thompson Sampling $Q^\mathrm{TS}$.
    \item As shown in the left panel of Figure \ref{figure_intro}, $Q^\mathrm{R2}$ also achieves substantially lower cumulative regret than $Q^\mathrm{TS}$, empirically confirming the faithfulness of our stationarization.
    \item Through the lens of online optimization, the under-performance of Thompson Sampling stems from the fact that its regularizer $\nu^\mathrm{TS}(\pi_t)$ deviates from the Bellman-optimal one $\nu^\mathrm{R2}(\pi_t)$, both conceptually (uncertainty vs. tension) and numerically (see the right panel of Figure \ref{figure_intro}).
    \item Guided by Bellman's principle, Thompson Sampling (and any other $\mathcal{R}^2$-finite policy) can now be improved through a standard policy-improvement step
    \[
    q_t^{\mathrm{TS}'}=\underset{q_t}{\mathrm{argmin}}\sbk{r^2(q_t;\pi_t)+q_t\cdot\E_{\pi_t}\sbk{V^\mathrm{TS}(\pi_{t+1})|A_t=\cdot}},
    \]
    where $V^\mathrm{TS}(\pi_{t+1})=\mathcal{R}^2(Q^\mathrm{TS};\pi_{t+1})$ is obtained by policy evaluation. Quite remarkably, a single policy-improvement step closes a substantial portion of the performance gap between Thompson Sampling and the $\mathcal{R}^2$-optimal policy, as illustrated later in the paper.
\end{itemize}

The rest of the paper is organized as follows: In Section \ref{sec_pre}, we review the Bayesian MAB problem. In Section \ref{sec_faith}, we develop the concept of faithful stationarization in detail. In Section \ref{sec_online}, we rediscover Thompson Sampling as an online optimization algorithm and compare its regularization mechanism with that of the Bellman-optimal policy. In Section \ref{sec_tvb}, we approximately implement the Bellman-optimal policy and benchmark Thompson Sampling through numerical experiments. In Section \ref{sec_pi}, we apply (principled) policy improvement to Thompson Sampling.

\section{Preliminaries}
\label{sec_pre}
\subsection{Bayesian stochastic bandits}
To begin, we recall the mechanism of a two-armed Bayesian stochastic bandit. The two arms are labeled with $1$ and $2$. Their joint reward distribution $P_\theta$ depends on an (unknown) environment parameter $\theta\in\Theta$. Before the game begins, $\theta$ is drawn from a prior distribution $\pi_0$ and remains fixed throughout the game. At each round, a potential reward vector is drawn independently from $P_\theta$, but only the entry corresponding to the pulled arm is observed. Conditional on $\theta$, these potential reward vectors form an independent and identically distributed (iid) sequence
\[
R_0|\theta,R_1|\theta,\dots\stackrel{\mathrm{iid}}{\sim}P_\theta,\;\;\theta\sim\pi_0.
\]
After $t$ rounds, each involving a partial observation of a potential reward vector, the posterior distribution $\pi_t$ of $\theta$ is obtained by updating $\pi_0$ according to Bayes' rule. For simplicity, we take the environment parameter to be the mean reward vector
\[
\E[R_0|\theta]=\E[(R_{1,0},R_{2,0})|\theta]=(\theta_1,\theta_2)=\theta.
\]
To make the next decision, Thompson Sampling draws $\theta'$ from $\pi_t$ and selects arm $A_t=\mathrm{argmax}(\theta_1',\theta_2')$ as if $\theta'$ were the true mean reward vector. After pulling the selected arm, based on the reward observed $R_{A_t,t}$, the belief is updated from $\pi_t$ to $\pi_{t+1}$, and the process repeats.

\subsection{An MDP view}
The Bayesian stochastic bandit can be viewed as an MDP \citep{gittins1979bandit}; see \citet{ghavamzadeh2015bayesian} for an illustrative example. The MDP formulation is as follows:
\begin{itemize}
    \item State: current belief $\pi_t$.
    \item Action: arm pulled $A_t$.
    \item Transition: updating $\pi_t$ to $\pi_{t+1}$ after observing the $A_t$-th entry of
    $R_t\sim P_{\theta''}$ where $\theta''\sim\pi_t$.
    \item MDP reward: expected next-round reward $\E_{\pi_t}R_{A_t,t}$.
\end{itemize}
Note that the next potential reward vector is drawn from the posterior predictive distribution ($R_t\sim P_{\theta''}$, $\theta''\sim\pi_t$), so the system can evolve forward without knowing which $\theta$ was drawn and fixed at the beginning. This MDP view provides a natural framework for analyzing Bayesian bandit algorithms directly, without resorting to frequentist analysis followed by integration over the prior. We adopt this view in our analysis, and our analysis is entirely Bayesian.

\begin{remark}[Justifying the MDP view]
    The MDP view is fully aligned with the original bandit mechanism (iid $R_t$ conditional on $\theta$). To see this, note that the posterior distribution $\pi_t$ fully characterizes the distribution of the never observed $\theta$ given the first $t$ observed rewards, and therefore the posterior predictive distribution fully characterizes the distribution of the next reward.
\end{remark}

Viewing the Bayesian stochastic bandit as an MDP, Thompson Sampling induces a Markov chain on the space of beliefs, as its transition from $\pi_t$ to $\pi_{t+1}$ only depends on a sample from the posterior ($\theta'\sim\pi_t$) and a sample from the posterior predictive ($R_t\sim P_{\theta''}$, $\theta''\sim\pi_t$); see Algorithm \ref{ts_gaussian_alg} (Gaussian rewards with Gaussian prior and posterior) and Algorithm \ref{ts_bernoulli_alg} (Bernoulli rewards with Beta prior and posterior).
Note that, unlike algorithms such as the Gittins index, Thompson Sampling does not attempt to solve the MDP in any dynamic programming sense. Despite having access to the posterior distribution that encapsulates all available information about the environment, Thompson Sampling merely draws a single sample and acts greedily with respect to it.

\begin{figure*}[ht]
\centering
\begin{minipage}[t]{0.49\textwidth}
\begin{algorithm}[H]
\caption{Thompson Sampling (Gaussian)}\label{ts_gaussian_alg}
\begin{algorithmic}
\State {\bf Initialize}: \(N(\mu_1, \sigma_1^2),\;N(\mu_2, \sigma_2^2),\;\tau^2,\;T\)
\For{\( t = 1, 2, \dots, T \)}
    \State Sample \[ (\theta'_1,\theta'_2) \sim N(\mu_1, \sigma_1^2)\times N(\mu_2, \sigma_2^2) \]
    \State Select \( A = \mathrm{argmax}(\theta'_1,\theta'_2) \)
    \State Observe \( R \sim N(\mu_A, \sigma^2_A+\tau^2)\)
    \State Update
    \State \( \mu_A\leftarrow (\mu_A/\sigma^2_A + R/\tau^2)/(1/\sigma^2_A + 1/\tau^2) \)
    \State \( \sigma^2_A \leftarrow 1/(1/\sigma^2_A + 1/\tau^2) \)
\EndFor
\end{algorithmic}
\end{algorithm}
\end{minipage}
\hfill
\begin{minipage}[t]{0.49\textwidth}
\begin{algorithm}[H]
\caption{Thompson Sampling (Bernoulli)}\label{ts_bernoulli_alg}
\begin{algorithmic}
\State {\bf Initialize}: \(\mathrm{Beta}(\alpha_1, \beta_1),\;\mathrm{Beta}(\alpha_2, \beta_2),\;T \)\vphantom{$\sigma_k^2(0)$}
\For{\( t = 1, 2, \dots, T\)}
    \State Sample \[ (\theta'_1,\theta'_2) \sim \mathrm{Beta}(\alpha_1, \beta_1)\times\mathrm{Beta}(\alpha_2, \beta_2) \]
    \State Select \( A = \mathrm{argmax}(\theta'_1,\theta'_2) \)
    \State Observe \( R \sim \mathrm{Ber}(\alpha_A/(\alpha_A+\beta_A))\)\vphantom{$\sigma_A^2$}
    \State Update
    \State \( \alpha_A \leftarrow \alpha_A + R \)\vphantom{$\mu_A/\sigma^2_A$}
    \State \( \beta_A \leftarrow \beta_A + (1 - R) \)\vphantom{$1/\sigma^2_A$}
\EndFor
\end{algorithmic}
\end{algorithm}
\end{minipage}
\end{figure*}

\section{Faithful Stationarization}
\label{sec_faith}
\subsection{Squared regret}
The Gittins index policy \citep{gittins1979bandit} harnesses the MDP formulation with the objective of maximizing cumulative discounted rewards \eqref{eqn_discount}, hence employing a common notion of stationarizing control problems.    Rothschild (1974) shows that, with probability one, the optimal policy eventually settles on a single arm (Theorem II), and that with positive probability this arm is not the optimal one (Theorem I). This behavior, often referred to as ``incomplete learning,"  stands in contrast to Robbins' principle that any policy aiming to maximize long-run average reward must pull all arms infinitely often \citep{robbins1952some}. Specifically, all policies  that  allocate a vanishing fraction of pulls to suboptimal arms are in an equivalence class that achieves long run average optimality.  
Of course this is a rather coarse notion of optimality and to inject  additional discriminatory power within this class, algorithm design in the bandit literature has focused on minimizing  cumulative regret \eqref{eqn_regret} across {\it finite} horizons $\{\mathcal{R}_T(Q;\pi_0):T\geq0\}$. This is clearly a sequence of non-stationary objectives, as optimal decisions depend on the remaining time in the problem horizon. 

In practice, the exact horizon is often unknown or indefinite, making horizon-dependent policies less appealing. More importantly, although finite-horizon problems admit Bellman recursions, their time dependence reduces computational tractability and obscures structural insight relative to stationary formulations. These considerations motivate the search for a stationary formulation (not the infinite horizon discounted one) that preserves the essence of long-term regret minimization (i.e., minimizing $\{\mathcal{R}_T(Q;\pi_0):T\geq0\}$), which we refer to as {\it faithful stationarization}.
\begin{remark}[Faithful stationarization]
    To be more explicit, by faithful stationarization we mean finding a stationary objective that remains tied to the original finite-horizon objective sequence, in the sense that controlling the former provides control over the latter.
\end{remark}
Recall that
\[
\begin{aligned}
    \mathcal{R}_T(Q;\pi_0)=&\E_{\pi_0}\sbk{\sum_{t=0}^{T-1}\sbk{\max(\theta_1,\theta_2)-\theta_{A_t}}}\\
    =&\E_{\pi_0}\sbk{\sum_{t=0}^{T-1}\E_{\pi_0}\sbk{\max(\theta_1,\theta_2)-\theta_{A_t}\Big|\pi_t}}\\
    =&\E_{\pi_0}\sbk{\sum_{t=0}^{T-1}r(q_t;\pi_t)},
\end{aligned}
\]
where $q_t$ is the distribution of $A_t|\pi_t$ under policy $Q$, and $r(q_t;\pi_t)=\E_{\pi_t}\sbk{\mathrm{max}(\theta_1,\theta_2)-\theta_{A_t}}$ is the expected next-round regret (given $\pi_t$).
To amalgamate  the cumulative regret sequence into a single quantity, we derive the simple regret bound \eqref{eqn_bound} mentioned in the introduction
\[
\begin{aligned}
    \mathcal{R}_T(Q;\pi_0)\leq&\E_{\pi_0}\sbk{\prs{\sum_{t=0}^{T-1}1}^{1/2}\prs{\sum_{t=0}^{T-1}r^2(q_t;\pi_t)}^{1/2}}\\
    \leq&\sqrt{T}\cdot\prs{\E_{\pi_0}\sbk{\sum_{t=0}^{T-1}r^2(q_t;\pi_t)}}^{1/2}\\
    \leq&\sqrt{\mathcal{R}^2(Q;\pi_0)\cdot T},
\end{aligned}
\]
where Cauchy's inequality and Jensen's inequality are used. As the leading constant in this regret bound, the (infinite-horizon cumulative) squared regret
\[
\mathcal{R}^2(Q;\pi_0)=\E_{\pi_0}\sbk{\sum_{t=0}^{\infty}r^2(q_t;\pi_t)}
\]
is a natural and meaningful objective to minimize.
In particular, Thompson Sampling $Q^{\mathrm{TS}}$ achieves finite squared regret, which is a direct corollary of the information-theoretic analysis in \cite{russo2016information}.
\begin{proposition}[$\mathcal{R}^2$-finiteness of $Q^\mathrm{TS}$]
\label{prop_ts_finite}
If there exists a finite constant $\sigma>0$ such that the posterior predictive distribution of the reward ($R_t\sim P_{\theta''}$, $\theta''\sim\pi_t$) is always $\sigma$-sub-Gaussian, then Thompson Sampling satisfies $\mathcal{R}^2(Q^{\mathrm{TS}};\pi_0)<\infty$, and hence $\mathcal{R}_T(Q^{\mathrm{TS}};\pi_0)=O(\sqrt{T})$.
\end{proposition}
This shows that the set of $\mathcal{R}^2$-finite policies is non-empty. Thanks to the regret bound \eqref{eqn_bound}, each policy in this set achieves $O(\sqrt{T})$ cumulative regret, which is minimax optimal \citep{auer2002nonstochastic}. These observations make the $\mathcal{R}^2$-optimal policy $Q^{\mathrm{R2}}=\mathrm{argmin}_Q\mathcal{R}^2(Q;\cdot)$ worth studying, as it not only achieves $O(\sqrt{T})$ cumulative regret but also attains the best possible constant in the regret bound \eqref{eqn_bound}. More importantly, studying the $\mathcal{R}^2$-optimal policy may shed light on the optimization considerations underlying Thompson Sampling, as $Q^{\mathrm{R2}}$ achieves what $Q^\mathrm{TS}$ achieves ($\mathcal{R}^2$-finiteness) but entirely through principled optimization (i.e., dynamic programming).

\begin{remark}[Regret bound minimization]
    Although $Q^\mathrm{R2}$ attains the best possible constant in the regret bound \eqref{eqn_bound}, it is optimal only in terms of minimizing this bound. In particular, it is not guaranteed that $Q^\mathrm{R2}$ is optimal in terms of minimizing $\{\mathcal{R}_T(Q;\pi_0):T\geq0\}$. In Section \ref{sec_tvb}, it is empirically illustrated that $Q^\mathrm{R2}$ significantly outperforms $Q^\mathrm{TS}$ in terms of minimizing $\{\mathcal{R}_T(Q;\pi_0):T\geq0\}$.
\end{remark}

\begin{remark}[Why square?]
    By replacing Cauchy's inequality with H\"older's inequality, the regret bound \eqref{eqn_bound} becomes
    \[
        \mathcal{R}_T(Q;\pi_0)\leq T^{1-1/p}\cdot\sbk{\mathcal{R}^p(Q;\pi_0)}^{1/p},
    \]
    where $\mathcal{R}^p(Q;\pi_0)$ denotes the $p$-th power analogue of squared regret for $p>1$. The choice $p=2$ is distinguished. When $p>2$, the regret bound grows faster than $\sqrt{T}$, which is not sharp. When $p<2$, the regret bound grows slower than $\sqrt{T}$, which would contradict the minimax optimality of $O(\sqrt{T})$ cumulative regret unless $\mathcal{R}^p(Q;\pi_0)=\infty$ for some $\pi_0$. Finally, when $p=2$, the regret bound \eqref{eqn_bound} is sharp, and there exists $Q$ (e.g., $Q^\mathrm{TS}$) such that $\mathcal{R}^2(Q;\pi_0)<\infty$ for all $\pi_0$ (Proposition \ref{prop_ts_finite}).
\end{remark}

\subsection{Bellman equation}
Squared regret minimization is a stationary formulation of the Bayesian MAB problem that remains faithful to the original goal of minimizing cumulative regret across finite horizons, as ensured by the regret bound \eqref{eqn_bound}. Since each $\mathcal{R}^2$-finite policy achieves sublinear cumulative regret, the new formulation is also faithful to Robbins' principle. More importantly, $O(\sqrt{T})$ cumulative regret can now be achieved not only by Thompson's heuristic but also by Bellman's principle. The corresponding stationary Bellman equation is
\begin{equation}
\label{eqn_Bellman}
V(\pi_t)=\min_{q_t}\sbk{r^2(q_t;\pi_t)+q_t\cdot\E_{\pi_t}\sbk{V(\pi_{t+1})|A_t=\cdot}},
\end{equation}
where $V(\pi_t)=\mathcal{R}^2(Q^\mathrm{R2};\pi_t)$ is the minimal squared regret incurred from $\pi_t$ onward, achieved by the $\mathcal{R}^2$-optimal policy. Basically, the current $V$-value equals the instantaneous squared regret plus the expected future $V$-value, minimized over all possible values of the pulling probability vector $q_t$.
Recall that the Bellman equation corresponding to the Gittins index policy is
\begin{equation}
\label{eqn_Bellman_dis}
V(\pi_t)=\max_{q_t}\sbk{q_t\cdot\E_{\pi_t}\theta+\gamma q_t\cdot\E_{\pi_t}\sbk{V(\pi_{t+1})|A_t=\cdot}},
\end{equation}
which maximizes cumulative discounted reward ($\gamma\in(0,1)$). The two Bellman equations highlight a structural difference between discounted reward maximization and squared regret minimization.

The maximization in \eqref{eqn_Bellman_dis} is linear in $q_t=(q_{1,t},q_{2,t})$, so the maximizer is either $(1,0)$ or $(0,1)$, i.e., $q_{1,t}\in\{0,1\}$. Consequently, the optimal policy corresponding to \eqref{eqn_Bellman_dis} is deterministic: at each round, one arm (the one with the highest Gittins index) is pulled with probability one. In contrast, the minimization in \eqref{eqn_Bellman} is quadratic in $q_t$ as 
\[
r^2(q_t;\pi_t)=\prs{\E_{\pi_t}\mathrm{max}(\theta_1,\theta_2)-q_t\cdot\E_{\pi_t}\theta}^2,
\]
so the minimizer ranges ``between'' $(1,0)$ and $(0,1)$, i.e., $q_{1,t}\in[0,1]$. Consequently, the optimal policy corresponding to \eqref{eqn_Bellman} is randomized: at each round, one arm is sampled and then pulled. Remarkably, both Thompson's heuristic and Bellman's principle intersect at the concept of {\it sampling} (randomization). Thompson Sampling successfully achieves $O(\sqrt{T})$ cumulative regret, while ``Bellman Sampling'' (the $\mathcal{R}^2$-optimal policy) achieves the same rate with the best possible constant $\sqrt{V(\pi_0)}$ in the regret bound \eqref{eqn_bound}. As we show next, these two sampling schemes share a common online optimization form, thereby revealing a deeper structural connection.

\section{Online Optimization Form}
\label{sec_online}
\subsection{The $\mathcal{R}^2$-optimal policy}
According to the Bellman equation \eqref{eqn_Bellman}, the $\mathcal{R}^2$-optimal policy is given by
\begin{equation}
\label{eqn_q_form_r2}
q_t^\mathrm{R2}=\underset{q_t}{\mathrm{argmin}}\sbk{r^2(q_t;\pi_t)+q_t\cdot\E_{\pi_t}\sbk{V(\pi_{t+1})|A_t=\cdot}}.
\end{equation}
In the two-armed case, the minimization is one-dimensional ($q_{1,t}+q_{2,t}=1$). However, parameterizing the decision by either $q_{1,t}$ or $q_{2,t}$ obscures the exploration-exploitation interpretation. In the frequentist setting, arm 1 can be fixed as the optimal arm, so $q_{1,t}$ and $q_{2,t}$ correspond to exploiting the optimal arm and exploring the suboptimal arm, respectively. In the Bayesian setting, however, the identity of the leading arm (determined by which posterior mean reward is currently higher) cannot be fixed, so neither $q_{1,t}$ nor $q_{2,t}$ has a fixed interpretation as exploitation or exploration. To resolve this issue, we shift focus to  the expected next-round reward
\[
x_t=q_t\cdot\E_{\pi_t}\theta=q_{1,t}\E_{\pi_t}\theta_1+q_{2,t}\E_{\pi_t}\theta_2
\]
as the decision variable, which lies between the two posterior mean rewards. A larger (respectively, smaller) value of $x_t$ always corresponds to more (respectively, less) exploitation, regardless of which arm is currently leading.
After this change of variable, the $\mathcal{R}^2$-optimal policy reveals an intuitive online optimization form
\begin{equation}
\label{eqn_form_r2}
x_t^\mathrm{R2}=\underset{x_t}{\mathrm{argmin}}\sbk{\prs{\E_{\pi_t}\max(\theta_1,\theta_2)-x_t}^2+\nu^\mathrm{R2}(\pi_t)x_t},
\end{equation}
where the {\it regularizer} is given by
\begin{equation}
\label{eqn_reg_r2}
\nu^\mathrm{R2}(\pi_t)=\sbk{\frac{\E_{\pi_t}\sbk{V(\pi_{t+1})|A_t=1}-\E_{\pi_t}\sbk{V(\pi_{t+1})|A_t=2}}{\E_{\pi_t}\theta_1-\E_{\pi_t}\theta_2}}_+. 
\end{equation}
The online objective in \eqref{eqn_form_r2} consists of two terms: an instantaneous loss term for exploitation (decreasing in $x_t$) and a linear regularization term for exploration (increasing in $x_t$). In this manner, the ``greediness" that would result from minimizing the first term alone is ``regularized" by the second term.

\begin{remark}[Positive part]
    After the change of variables, the regularizer in \eqref{eqn_form_r2} should equal the ratio in \eqref{eqn_reg_r2}. We retain only its positive part in the definition of $\nu^\mathrm{R2}(\pi_t)$, because it is clear that $x_t^\mathrm{R2}=\max(\E_{\pi_t}\theta_1,\E_{\pi_t}\theta_2)$ whenever the coefficient of the regularization term is negative.
\end{remark}

\begin{remark}[Equal means]
    When $\E_{\pi_t}\theta_1=\E_{\pi_t}\theta_2$, the range of $x_t$ collapses to a single point. Nevertheless, $q_t^\mathrm{R2}$ can still be recovered from $x_t^\mathrm{R2}$ by imagining an infinitesimal difference between the two posterior mean rewards (e.g., translating the distribution of $\theta_1$ by $\epsilon$ and letting $\epsilon\rightarrow0$). In this tied case, the instantaneous squared regret $r^2(q_t;\pi_t)$ is constant in $q_t$. As a result, the $\mathcal{R}^2$-optimal policy $q_t^\mathrm{R2}$ given by \eqref{eqn_q_form_r2} places all its mass on the arm that yields lower expected future $V$-value.
\end{remark}

\begin{remark}[Dimensionality]
In the two-armed case, the decision variable $q_t$ is effectively one-dimensional ($q_{1,t}+q_{2,t}=1$), so vector $q_t^{\mathrm{R2}}$ can be recovered from scalar $x_t^{\mathrm{R2}}$. When there are more than two arms, somewhat surprisingly, the vector $q_t^{\mathrm{R2}}$ can still be recovered from the scalar $x_t^{\mathrm{R2}}$, as discussed in Section \ref{sec_pi}.
\end{remark}

\subsection{Tension measure}
In the online optimization form \eqref{eqn_form_r2}, the regularizer $\nu^\mathrm{R2}(\pi_t)$ determines how much exploitation (i.e., increasing $x_t$ to reduce the first term) should be traded off against exploration (i.e., decreasing $x_t$ to reduce the second term), quantifying the current tension between exploration and exploitation. To be specific, in \eqref{eqn_reg_r2}, the first (respectively, second) term in the numerator represents the future squared regret after pulling arm 1 (respectively, arm 2). If the numerator is positive, then arm 2 favors exploration, as pulling it leads to lower future squared regret. If the denominator is positive, then arm 1 favors exploitation, as pulling it leads to higher immediate mean reward. As a result, when the ratio is positive, there is a clear tension between exploration and exploitation. Quantifying this tension requires looking into the ``far future," as computing $\nu^\mathrm{R2}(\pi_t)$ (to implement $Q^\mathrm{R2}$) requires solving the Bellman equation \eqref{eqn_Bellman}.

Is there a tension measure that avoids solving the Bellman equation \eqref{eqn_Bellman}? There is one hidden in Information-Directed Sampling (IDS) \citep{russo2014learningIDS}, another notable member of the family of $\mathcal{R}^2$-finite policies.
Recall that IDS is given by
\begin{equation}
\label{eqn_ids}
q_t^\mathrm{IDS}=\underset{q_t}{\mathrm{argmin}}\frac{r^2(q_t;\pi_t)}{\mathcal{I}(q_t;\pi_t)},
\end{equation}
where $\mathcal{I}(q_t;\pi_t)$ is the ``information gain'' from executing $q_t$ (defined later). It is straightforward to rewrite IDS in the online optimization form \eqref{eqn_form_r2}, and the corresponding regularizer is
\begin{equation}
\label{eqn_ids_reg}
\nu^\mathrm{IDS}(\pi_t)=\sbk{\frac{\mathcal{I}((0,1);\pi_t)-\mathcal{I}((1,0);\pi_t)}{\E_{\pi_t}\theta_1-\E_{\pi_t}\theta_2}}_+\cdot\min_{q_t}\frac{r^2(q_t;\pi_t)}{\mathcal{I}(q_t;\pi_t)}.
\end{equation}
The second term is the minimal ``information ratio,'' while the first term, similar to $\nu^\mathrm{R2}(\pi_t)$ in \eqref{eqn_reg_r2}, is a tension measure. The ratio is positive when one arm gives more reward while the other arm gives more information (i.e., when there is a clear tension between exploration and exploitation).

\subsection{Thompson Sampling}
Rather than just focusing on  the $\mathcal{R}^2$-optimal policy itself, we now shift attention to the  online optimization form \eqref{eqn_form_r2} associated with it.  In particular, this form reveals what a reasonable bandit algorithm (that achieves $O(\sqrt{T})$ cumulative regret) should look like when the MAB problem is viewed through the lens of online optimization: minimizing the instantaneous squared regret with some linear regularization. We now show that Thompson Sampling also takes this form, focusing on the two-armed case
\[
q_t^{\mathrm{TS}}=(P_{\pi_t}(\theta_1>\theta_2),P_{\pi_t}(\theta_1\leq\theta_2)).
\]
The corresponding expected next-round reward is
\begin{equation}
\label{eqn_ts_mean}
x_t^\mathrm{TS}=P_{\pi_t}(\theta_1>\theta_2)\E_{\pi_t}\theta_1+P_{\pi_t}(\theta_1\leq\theta_2)\E_{\pi_t}\theta_2.
\end{equation}
In fact, for Thompson Sampling, the $K$-armed case can be viewed as repeating the two-armed case $K$ times to determine the $K$ pulling probabilities.
For example, the probability of pulling arm 1
\[
q_{1,t}^{\mathrm{TS}}=P_{\pi_t}(\theta_1>\theta_2,\dots,\theta_K)=P_{\pi_t}(\theta_1>\theta_{-1})
\] 
is determined by hedging between two random variables: $\theta_1$ and $\theta_{-1}=\max\{\theta_2,\dots,\theta_K\}$.
In this sense, Thompson's focus on {\it two samples} back in 1933 already contains the core idea of Thompson Sampling as a multi-armed bandit algorithm. This idea now admits an online optimization form.
\begin{theorem}[Online optimization]
\label{thm_ts}
The online optimization form of Thompson Sampling is
\[
x_t^{\mathrm{TS}}=\underset{x_t}{\mathrm{argmin}}\sbk{\prs{\E_{\pi_t}\max(\theta_1,\theta_2)-x_t}^2+\nu^\mathrm{TS}(\pi_t)x_t},
\]
where $\nu^{\mathrm{TS}}(\pi_t)=\mathrm{Cov}_{\pi_t}(\theta_1-\theta_2,\mathrm{sign}(\theta_1-\theta_2))$.
\end{theorem}
All proofs are in Section \ref{sec_proof}. According to Thompson's original idea, posterior sampling corresponds to the expected next-round reward $x_t^{\mathrm{TS}}$ given by \eqref{eqn_ts_mean}.
This $x_t^{\mathrm{TS}}$ now emerges from the above online optimization form, which is identical to \eqref{eqn_form_r2} except it uses a different regularizer.
To wit, the Bellman-optimal regularizer $\nu^\mathrm{R2}(\pi_t)$ is replaced by a simpler one $\nu^\mathrm{TS}(\pi_t)$, which turns out to be of covariance form. 

In this online optimization framework, different policies are fully characterized by their regularizers, so policy design becomes an exercise in ``regularizer engineering."
In particular, the randomized behavior of Thompson Sampling now admits a clean variational description, which allows us to improve it by modifying its regularizer.
A natural first attempt is to multiply the regularizer by a constant. Unfortunately, this modification gives rise to a failure known as {\it incomplete learning}.
\begin{proposition}[Incomplete learning]
\label{prop_incomplete}
For each $\lambda\neq1$, there exists a prior under which the policy
\[
x_t^{\lambda}=\underset{x_t}{\mathrm{argmin}}\sbk{\prs{\E_{\pi_t}\max(\theta_1,\theta_2)-x_t}^2+\lambda\nu^\mathrm{TS}(\pi_t)x_t}
\]
suffers from incomplete learning, i.e., it fully commits to one arm while the alternative may be optimal.
\end{proposition}
This result highlights that regularizer engineering is a delicate task. To make meaningful modifications that improve Thompson Sampling, we need principled guidance on what constitutes a better regularizer. In this regard, the Bellman-optimal regularizer $\nu^{\mathrm{R2}}(\pi_t)$ provides exactly such a compass, as $Q^\mathrm{R2}$ attains the best possible constant in the regret bound \eqref{eqn_bound}. By comparing the two regularizers $\nu^{\mathrm{TS}}(\pi_t)$ and $\nu^{\mathrm{R2}}(\pi_t)$ both analytically and numerically, we devote much of the sequel  to identifying and addressing the issues of Thompson Sampling in this principled manner. In particular, applying a standard policy-improvement step (Section \ref{sec_pi}) cures a fundamental issue of Thompson Sampling in the $K$-armed case, where ``repeating the two-armed case $K$ times'' may not work well. 

\subsection{Uncertainty measure}
Before diving into a comparison with the tension measure $\nu^\mathrm{R2}(\pi_t)$, we take a closer look at $\nu^\mathrm{TS}(\pi_t)$, which turns out to be an uncertainty measure. In Theorem \ref{thm_ts}, the regularizer of Thompson Sampling $\nu^\mathrm{TS}(\pi_t)$ is the covariance between the following two fundamental quantities:
\[
\begin{aligned}
    \Delta&=\theta_1-\theta_2\;\;\text{the reward gap between the two arms},\\
    \Lambda&=\mathrm{sign}(\theta_1-\theta_2)\;\;\text{the identity of the optimal arm}.
\end{aligned}
\]
The study of the relationship between a metric (continuous) variable and a dichotomous (binary) variable dates back to \cite{pearson1909new}, and their ``biserial'' covariance admits a well-known expression; see, e.g., \cite{lev1949point}.
\begin{proposition}[Covariance factorization]
\label{prop_cov}
If $\var_{\pi_t}\Lambda=0$, then $\cov_{\pi_t}(\Delta,\Lambda)=0$. Otherwise,
\[
\frac{\cov_{\pi_t}(\Delta,\Lambda)}{\var_{\pi_t}\Lambda}=\frac{\E_{\pi_t}[\Delta|\Delta>0]+\E_{\pi_t}[-\Delta|\Delta\leq0]}{2}.
\]
\end{proposition}
The covariance $\cov_{\pi_t}(\Delta,\Lambda)$ is the product of two terms: a variance and the average of two expectations. The variance of the identity of the optimal arm $\var_{\pi_t}\Lambda$ measures the remaining uncertainty about which arm is better. 
The two expectations $\E_{\pi_t}[\Delta|\Delta>0]$ and $\E_{\pi_t}[-\Delta|\Delta\leq0]$ correspond to the expected regret incurred by pulling arm 2 when arm 1 is optimal, and by pulling arm 1 when arm 2 is optimal.
Their average can be viewed as a notion of instantaneous regret.
Taken together, the regularizer of Thompson Sampling $\nu^\mathrm{TS}(\pi_t)$ is a regret-scaled uncertainty measure. In other words, Thompson Sampling measures uncertainty (via the biserial covariance) so as to guide exploration.

\begin{figure}[ht]
    \centering
    \begin{minipage}{0.49\textwidth}
        \centering
        \includegraphics[width=0.9\linewidth]{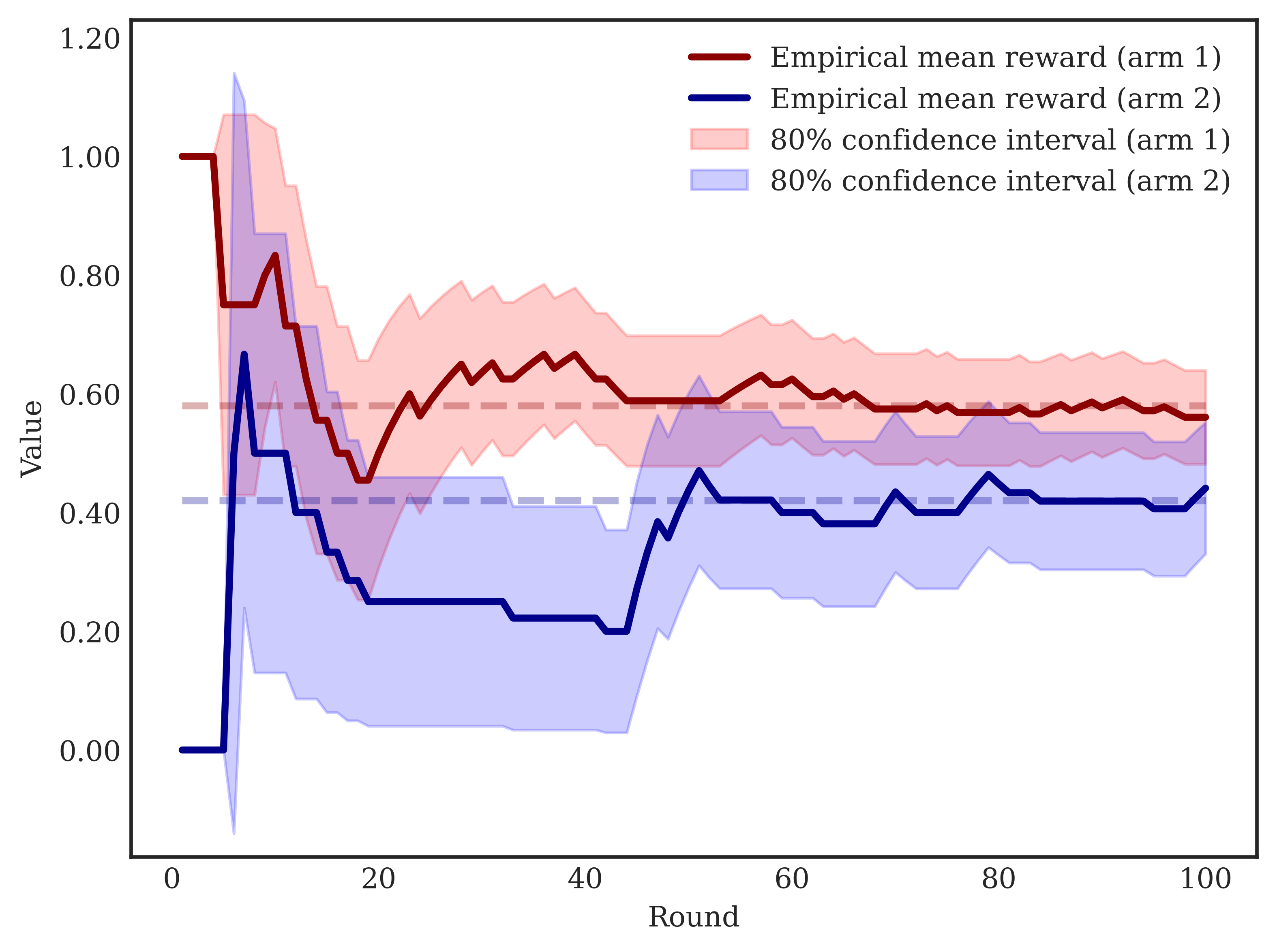}
    \end{minipage}
    \hfill
    \begin{minipage}{0.49\textwidth}
        \centering
        \includegraphics[width=0.9\linewidth]{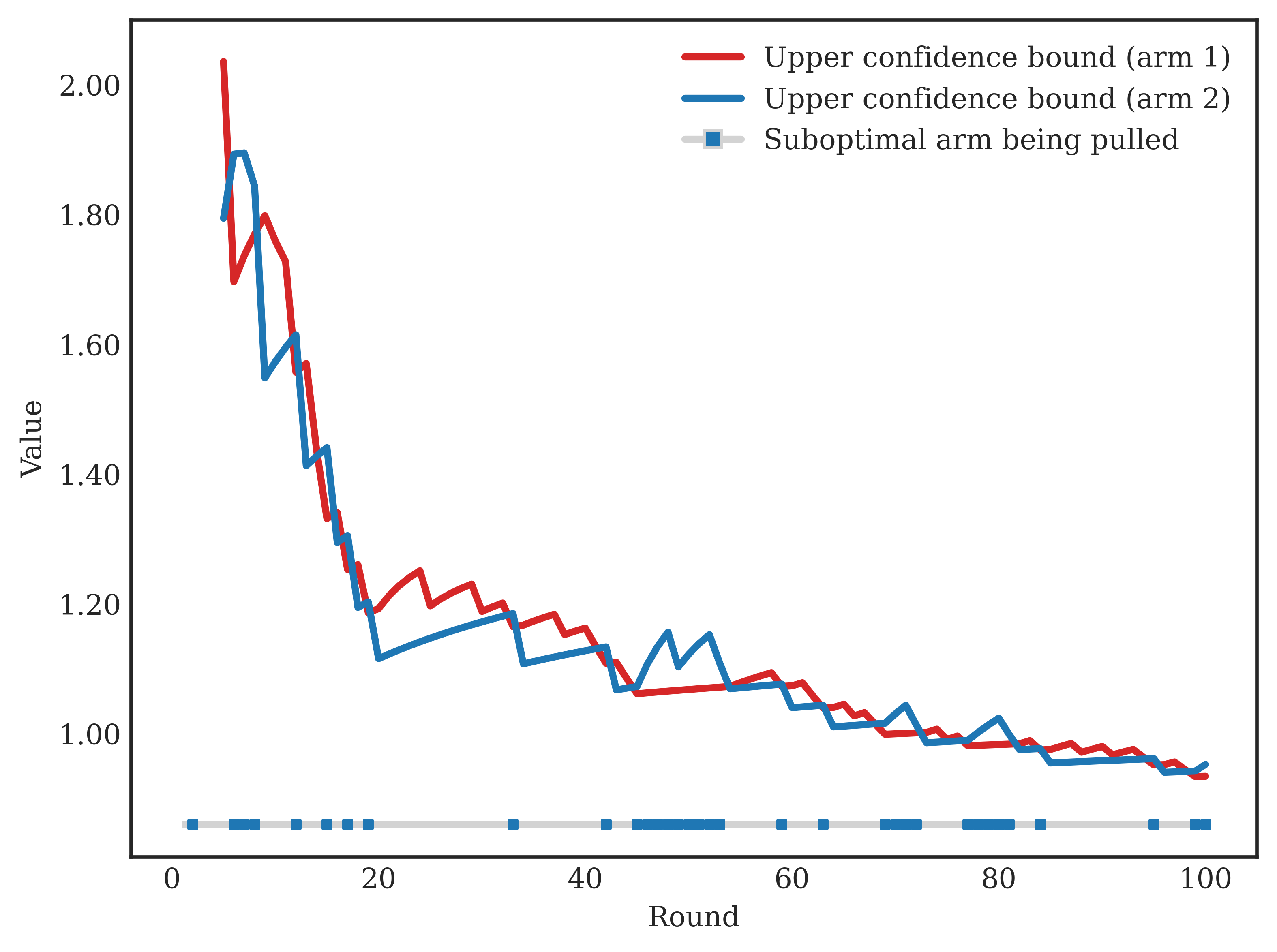}
    \end{minipage}

    \caption{
    UCB plays a two-armed Bernoulli bandit. Left: confidence intervals around empirical means. Right: upper confidence bounds. The suboptimal arm (arm 2) is pulled whenever the corresponding upper confidence bound (blue) is higher.
    }
    \label{figure_5_ucb}
\end{figure}

This new interpretation of Thompson Sampling naturally brings to mind the Upper Confidence Bound (UCB) algorithm \citep{auer2002finite}, which also measures uncertainty (via confidence intervals) to guide exploration. Recall that UCB, in the case of rewards that take value in the interval $[0,1]$,  is given by
\[
A_t=\underset{k}{\mathrm{argmax}}\prs{\hat{\mu}_{k,t}+\sqrt{\frac{2\log t}{N_{k,t}}}},
\]
where $N_{k,t}$ is the number of times arm $k$ has been pulled up to time $t$, and $\hat{\mu}_{k,t}$ is the corresponding empirical mean reward.
As suggested by the central limit theorem, the uncertainty of $\hat{\mu}_{k,t}$ can be represented by a confidence interval (left panel of Figure \ref{figure_5_ucb}), the width of which is of order $1/\sqrt{N_{k,t}}$. 
The two confidence intervals are then scaled by $\sqrt{2\log t}$, creating a catch-up game between the two upper confidence bounds (right panel of Figure \ref{figure_5_ucb}), which guides the exploration of UCB.

\begin{figure}[ht]
    \centering
    \begin{minipage}{0.49\textwidth}
        \centering
        \includegraphics[width=0.9\linewidth]{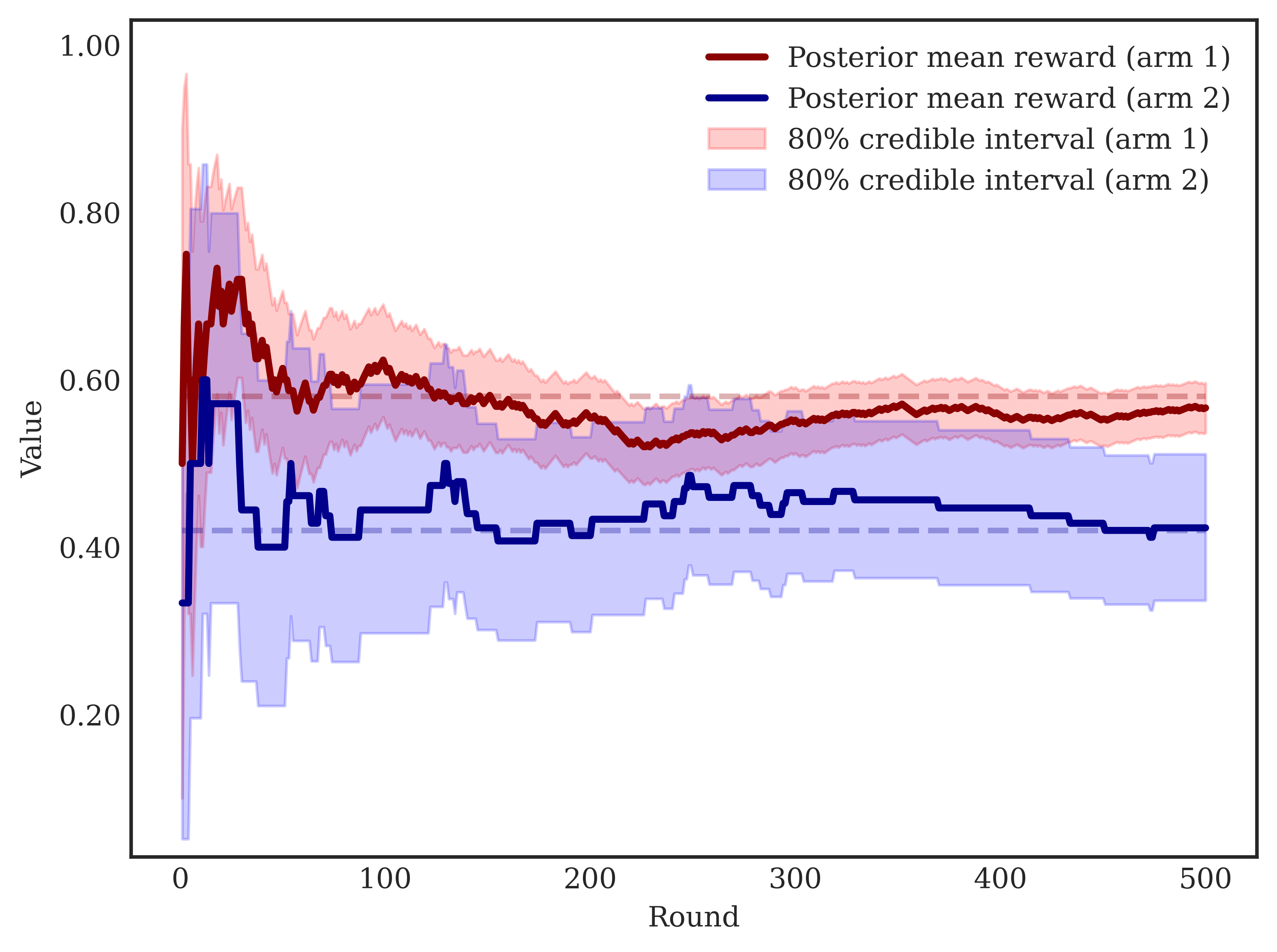}
    \end{minipage}
    \hfill
    \begin{minipage}{0.49\textwidth}
        \centering
        \includegraphics[width=0.9\linewidth]{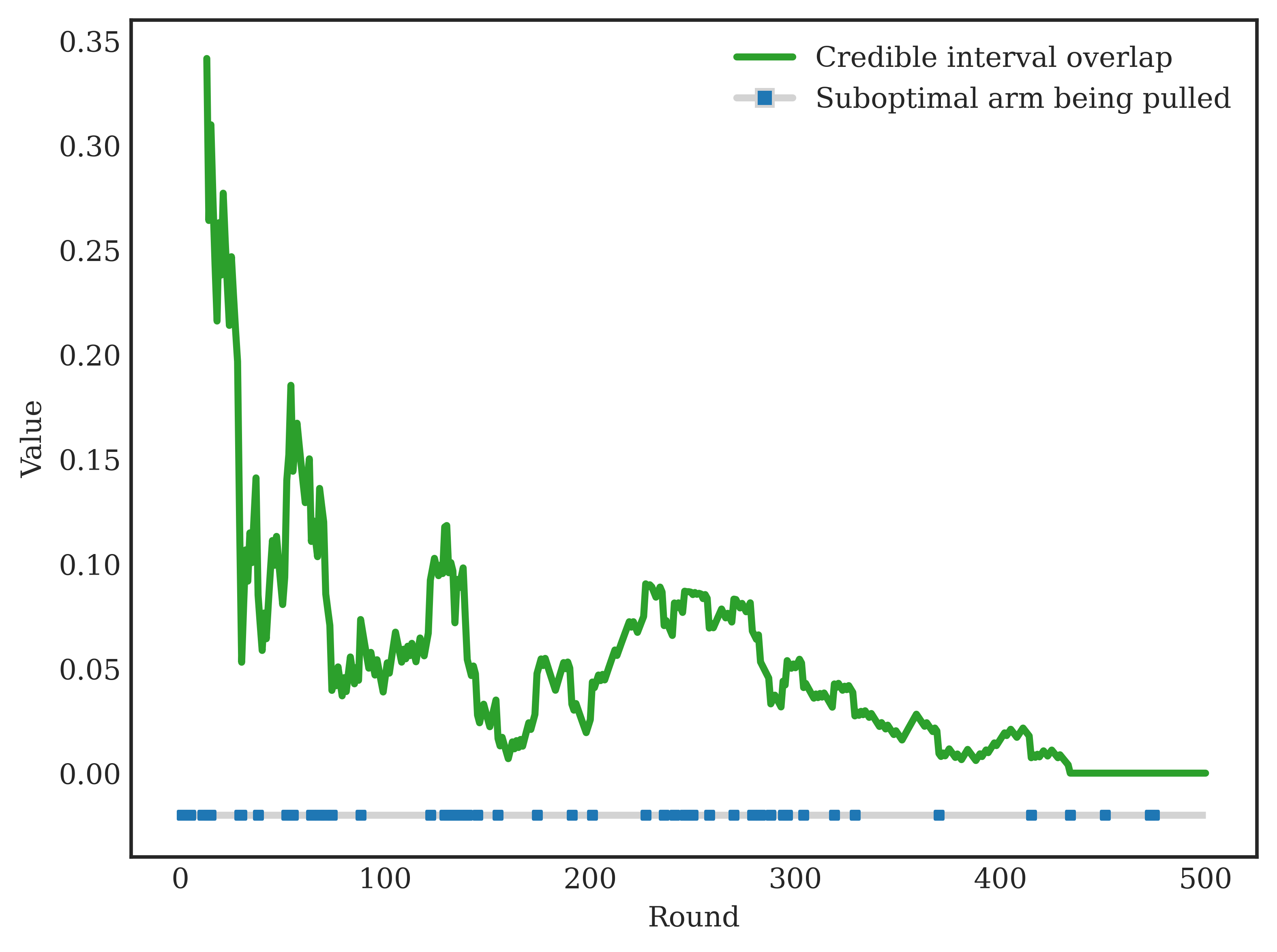}
    \end{minipage}

    \caption{
    Thompson Sampling plays a two-armed Bernoulli bandit. Left: credible intervals around posterior means. Right: the overlap of credible intervals. The overlap, when present, 
    reflects the frequency of pulling the suboptimal arm (arm 2).
    }
    \label{figure_5_ts}
\end{figure}

We now proceed to visualize how the exploration of Thompson Sampling is guided. According to its original definition, Thompson Sampling leverages uncertainty to regularize greediness in a probabilistic manner: the higher the probability that the current leader is not truly optimal, the more frequently the other arm is sampled. This probability is intuitively reflected by the overlap of {\it credible intervals}, the Bayesian analogue of confidence intervals in the frequentist domain\footnote{The 80\% credible interval of a distribution is $(F^{-1}(0.1),F^{-1}(0.9))$ where $F$ denotes the CDF.} (left panel of Figure \ref{figure_5_ts}). But note that despite  the credible intervals eventually detaching,  exploration does continue (right panel of Figure \ref{figure_5_ts}). This issue is resolved by $\cov_{\pi_t}(\Delta,\Lambda)$, the uncertainty measure that arises from the online optimization form of Thompson Sampling.

\begin{figure}[ht]
    \centering
    \begin{minipage}{0.49\textwidth}
        \centering
        \includegraphics[width=0.9\linewidth]{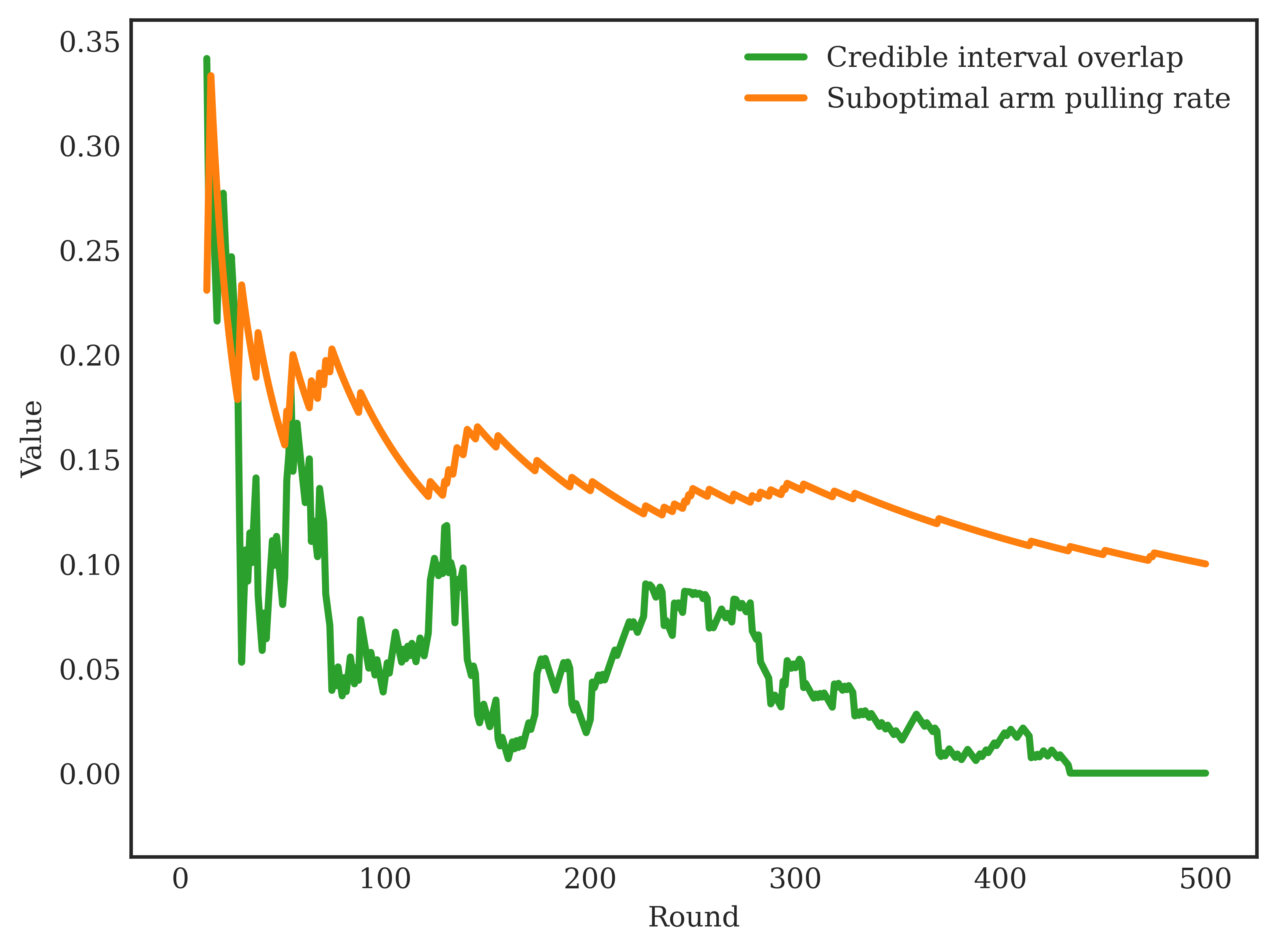}
    \end{minipage}
    \hfill
    \begin{minipage}{0.49\textwidth}
        \centering
        \includegraphics[width=0.9\linewidth]{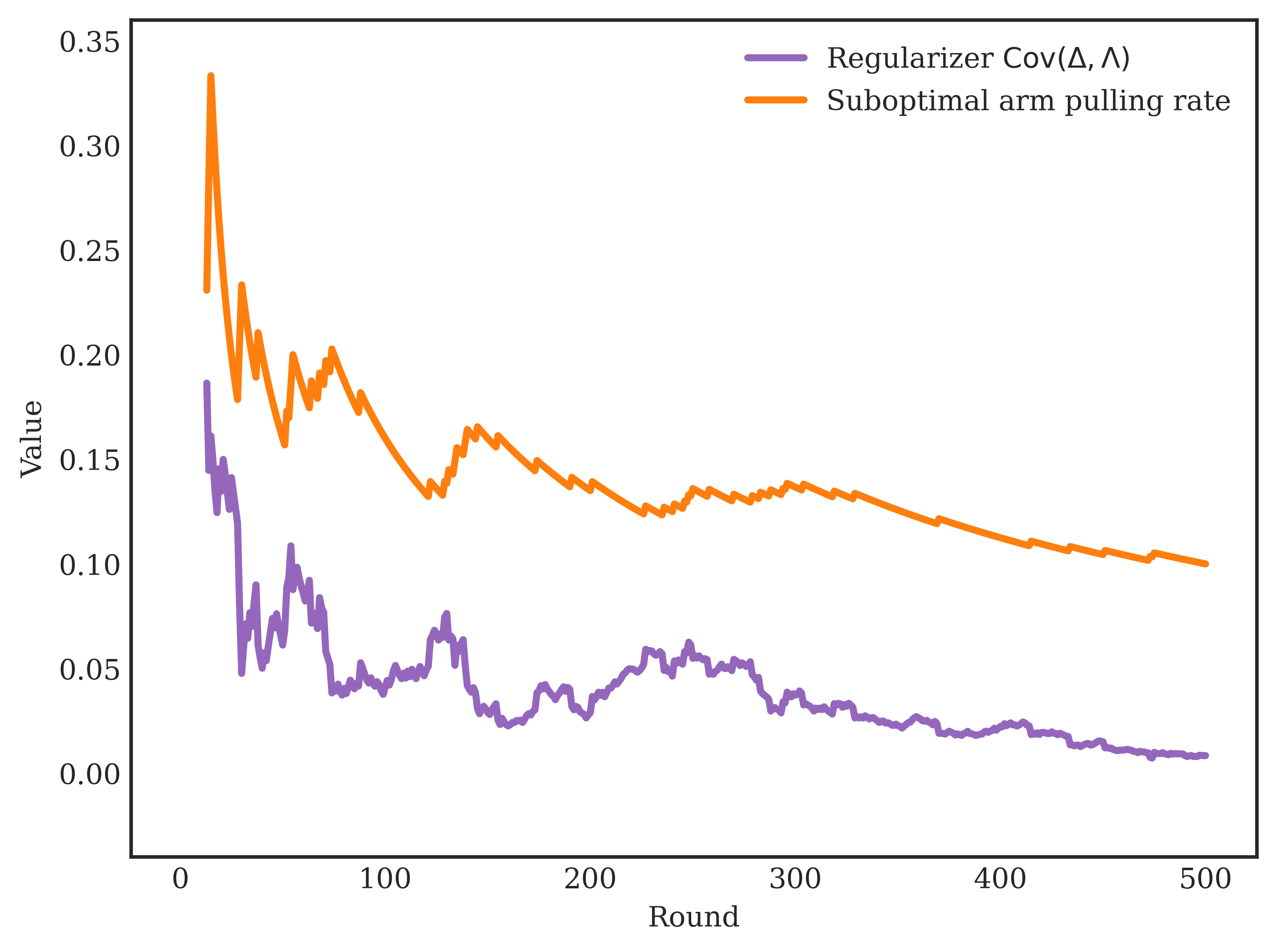}
    \end{minipage}

    \caption{
    Thompson Sampling plays a two-armed Bernoulli bandit. Left: the overlap of credible intervals vs. the pulling rate of the suboptimal arm. Right: the regularizer of Thompson Sampling vs. the pulling rate of the suboptimal arm.
    }
    \label{figure_5_cov}
\end{figure}

In Figure \ref{figure_5_cov}, we compare the rate at which the suboptimal arm is pulled (i.e., $N_{2,t}/t$), first with the overlap of credible intervals (left panel of Figure \ref{figure_5_cov}), and then with the biserial covariance (right panel of Figure \ref{figure_5_cov}). On the left, we see that the overlap, as an intuitive proxy for uncertainty, does guide the exploration to some extent, until it vanishes. The larger the overlap, the more Thompson Sampling allocates pulls to the suboptimal arm in order to resolve the uncertainty there. On the right, we see that the behavior of the ``exploration rate'' is captured indefinitely by the biserial covariance. This connection is not a coincidence: the biserial covariance, the quantitative notion of uncertainty, temporally correlates strongly (a Pearson correlation of 0.995) with the more qualitative notion of overlap of credible intervals.

\subsection{The two regularizers}
Recall that $\nu^\mathrm{R2}(\pi_t)$ measures the current tension between exploration and exploitation, so the $\mathcal{R}^2$-optimal policy pulls the runner-up arm more often when doing so yields greater future benefit. In contrast, $\nu^\mathrm{TS}(\pi_t)$ measures the remaining uncertainty about which arm is better, so Thompson Sampling pulls the runner-up arm more often when the identity of the optimal arm is less clear. This comparison reveals that the motivation behind Thompson Sampling pulling the runner-up arm is not entirely compelling, as doing so does not necessarily resolve uncertainty optimally, especially when the leading arm is more uncertain than the runner-up arm. For example, when $\pi_t=N(1,100)\times N(-1,0.01)$, arm 1 favors both exploration (learning more) and exploitation (earning more). In this case, there is uncertainty but little tension, i.e., $\nu^\mathrm{TS}(\pi_t)\gg\nu^\mathrm{R2}(\pi_t)$, so Thompson Sampling pulls arm 2 more frequently than necessary. The conceptual difference (uncertainty vs. tension) between the two regularizers motivates us to make further comparisons between the two algorithms, investigated in the next section. 

\section{Thompson vs. Bellman}
\label{sec_tvb}
To benchmark Thompson Sampling, we take a closer look at the $\mathcal{R}^2$-optimal policy, presenting a closed-form solution (to the Bellman equation \eqref{eqn_Bellman}) in the one-armed case and an approximate implementation in the two-armed case. Comparing the two regularizers allows us to identify and address the issues of Thompson Sampling, underscoring the appeal of a principled framework with a well-defined benchmark.

\subsection{One-armed Bandit}
In the one-armed case, where one of the two arms is fully known, the $\mathcal{R}^2$-optimal policy turns out to be fully tractable, i.e., the Bellman equation \eqref{eqn_Bellman} can be solved in closed form.
Without loss of generality, we take arm 2 to be the known arm, with $\theta_2\equiv0$ (i.e., $\theta_2\sim\delta_0$).
The Bellman equation \eqref{eqn_Bellman} becomes
\[
0=\min_{q_t}\sbk{\prs{\E_{\pi_t}[(\theta_1)_+]-q_{1,t}\E_{\pi_t}\theta_1}^2+q_{1,t}\prs{\E_{\pi_t}\sbk{V(\pi_{t+1})|A_t=1}-V(\pi_t)}},
\]
where $\E_{\pi_t}\sbk{V(\pi_{t+1})|A_t=2}-V(\pi_t)$ disappears as pulling the known arm 2 produces no posterior update ($\pi_{t+1}=\pi_t$). 
Since $\E_{\pi_t}\sbk{V(\pi_{t+1})|A_t=1}-V(\pi_t)\leq0$ is necessary for the minimum to be $0$,
the minimization is equivalent to
\[
V(\pi_t)-\E_{\pi_t}[V(\pi_{t+1})|A_t=1]=\min_{q_{1,t}}\sbk{\frac{\prs{\E_{\pi_t}[(\theta_1)_+]-q_{1,t}\E_{\pi_t}\theta_1}^2}{q_{1,t}}},
\]
where the minimizer $q_{1,t}^\mathrm{R2}$ can be computed in closed form.
\begin{proposition}[Closed-form solution]
\label{prop_one_arm}
When $\theta_2\equiv0$ and $\E_{\pi_t}\theta_1\neq0$, the $\mathcal{R}^2$-optimal policy pulls the unknown arm 1 with probability
\[
q^\mathrm{R2}_{1,t}=\min\prs{\frac{\E_{\pi_t}[(\theta_1)_+]}{|\E_{\pi_t}\theta_1|},1}.
\]
\end{proposition}

In Figure \ref{figure_intro} (at the end of the introduction), we compare Thompson Sampling with the $\mathcal{R}^2$-optimal policy in the one-armed case. The left panel of Figure \ref{figure_intro} shows that the $\mathcal{R}^2$-optimal policy achieves substantially lower cumulative regret than Thompson Sampling, confirming the faithfulness of our stationarization
(i.e., the benefit of minimizing the regret bound \eqref{eqn_bound}). 
The reason behind the gap can be understood via regularizer comparison, thanks to the shared online optimization form.
In the right panel of Figure \ref{figure_intro}, we compare the two regularizers along the prior sequence $\pi_0(\mu)=N(\mu,1)\times\delta_0$, with $\mu\in[-3,0]$, to illustrate how the two policies respond as the cost of exploration decreases. As the posterior mean reward of the unknown arm 1 increases from $-3$ to $0$, the cost of exploration decreases from $3$ to $0$. In particular, when $\mu=0$ (so the two arms have the same posterior mean reward), the exploration of arm~1 becomes cost-free and should therefore be carried out with probability one. The right panel of Figure \ref{figure_intro} shows that the regularizer of the $\mathcal{R}^2$-optimal policy does grow sharply, thereby encouraging pure exploration (i.e., pulling arm~1 with probability one).
In contrast, the regularizer of Thompson Sampling grows gradually, indicating that it does not fully account for the vanishing cost of exploration, as it is fundamentally an uncertainty measure.

Before turning to the two-armed case, we make an interesting observation about pure exploration. In Proposition \ref{prop_one_arm}, note that the condition $\E_{\pi_t}[(\theta_1)_+]/|\E_{\pi_t}\theta_1|=1$ marks a phase transition between randomized ($q^\mathrm{R2}_{1,t}<1$) and deterministic ($q^\mathrm{R2}_{1,t}=1$) behavior. In the Gaussian case (Algorithm 1), this phase transition can be characterized explicitly.
\begin{proposition}[Phase transition]
\label{prop_gaussian}
When $\theta_2\equiv0$ and $\theta_1\sim N(\mu_t,\sigma_t^2)$ under $\pi_t$,
\[
q^\mathrm{R2}_{1,t}=1\;\;\Leftrightarrow\;\;\mu_t/\sigma_t\geq\bar{x},
\]
where $\bar{x}\approx-0.276$ is the unique root of the increasing function $x\Phi(x)+\phi(x)+x$.
\end{proposition}
That is, the unknown arm 1 is pulled with probability one if and only if its signal-to-noise ratio exceeds $-0.276$. The value $0.276$ can be interpreted as the relative price Bellman is willing to pay for the exploratory benefit of pulling the unknown arm 1. When the cost of exploration falls below this threshold (e.g., $\mu_t=-0.2$, $\sigma_t=1$, and $-\mu_t/\sigma_t=0.2<0.276$), Bellman pulls the unknown arm 1 with probability one to fully capitalize on the available arbitrage. This quantitative result further illustrates the value of introducing Bellman's principle into the MAB problem through our squared regret formulation.

\begin{remark}[Phase transition]
If we look closely at the right panel of Figure \ref{figure_intro}, we can spot the phase transition of the $\mathcal{R}^2$-optimal policy at $-0.276$, where the regularizer curve becomes slightly less smooth than elsewhere.
\end{remark}

\subsection{Two-armed Bandit}
In the two-armed case, the Bellman equation \eqref{eqn_Bellman} can no longer be solved in closed form, but we are able to approximately implement the $\mathcal{R}^2$-optimal policy for Bernoulli bandits. 
In the Beta-Bernoulli setting (Algorithm 2), both the prior and posterior are Beta, so the belief space is parametrized by two pairs of positive integers
\[
\{\mathrm{Beta}(\alpha_1,\beta_1)\times\mathrm{Beta}(\alpha_2,\beta_2):\alpha_1,\beta_1,\alpha_2,\beta_2\geq1\}.
\]
Let 
$
V_{\alpha_1,\beta_1,\alpha_2,\beta_2}=V(\mathrm{Beta}(\alpha_1,\beta_1)\times\mathrm{Beta}(\alpha_2,\beta_2))
$
be the solution to the Bellman equation \eqref{eqn_Bellman}
\[
\begin{aligned}
    V_{\alpha_1,\beta_1,\alpha_2,\beta_2}=&\min_{p,q}\Big[\prs{\tilde{E}_{\alpha_1,\beta_1,\alpha_2,\beta_2}-(pE_{\alpha_1,\beta_1}+qE_{\alpha_2,\beta_2})}^2\\
    &\qquad\;\;+p(E_{\alpha_1,\beta_1}V_{\alpha_1',\beta_1,\alpha_2,\beta_2}+\bar{E}_{\alpha_1,\beta_1}V_{\alpha_1,\beta_1',\alpha_2,\beta_2})\\
    &\qquad\;\;+q(E_{\alpha_2,\beta_2}V_{\alpha_1,\beta_1,\alpha_2',\beta_2}+\bar{E}_{\alpha_2,\beta_2}V_{\alpha_1,\beta_1,\alpha_2,\beta_2'})\Big],
\end{aligned}
\]
where $p+q=1$, $\alpha_1'=\alpha_1+1$, $E_{\alpha_1,\beta_1}=\alpha_1/(\alpha_1+\beta_1)$, $\bar{E}_{\alpha_1,\beta_1}=1-E_{\alpha_1,\beta_1}$, and 
\[
\tilde{E}_{\alpha_1,\beta_1,\alpha_2,\beta_2}=\E\max(\mathrm{Beta}(\alpha_1,\beta_1),\mathrm{Beta}(\alpha_2,\beta_2)).
\]
Note that $V$ is symmetric (i.e., $V_{\alpha_1,\beta_1,\alpha_2,\beta_2}=V_{\alpha_2,\beta_2,\alpha_1,\beta_1}$), as swapping arm labels does not change the squared regret achieved by the $\mathcal{R}^2$-optimal policy.
The benefit of pulling arm 1 is
\[
V'_{\alpha_1,\beta_1,\alpha_2,\beta_2}=V_{\alpha_1,\beta_1,\alpha_2,\beta_2}-E_{\alpha_1,\beta_1}V_{\alpha_1',\beta_1,\alpha_2,\beta_2}-\bar{E}_{\alpha_1,\beta_1}V_{\alpha_1,\beta_1',\alpha_2,\beta_2}.
\]
By the symmetry of $V$, the benefit of pulling arm 2 is simply $V'_{\alpha_2,\beta_2,\alpha_1,\beta_1}$. To present the approximate implementation of the $\mathcal{R}^2$-optimal policy, we collect key properties of the benefit function $V'$ below.
\begin{proposition}[Benefit function]
\label{prop_benefit}
The benefit function $V'$ satisfies the following properties:
\begin{itemize}
    \item Backward recursion (part 1):
    \[
    \begin{aligned}
    &V'_{\alpha_1,\beta_1,\alpha_2,\beta_2}-E_{\alpha_2,\beta_2}V'_{\alpha_1,\beta_1,\alpha_2',\beta_2}-\bar{E}_{\alpha_2,\beta_2}V'
    _{\alpha_1,\beta_1,\alpha_2,\beta_2'}\\
    =&V'_{\alpha_2,\beta_2,\alpha_1,\beta_1}-E_{\alpha_1,\beta_1}V'_{\alpha_2,\beta_2,\alpha_1',\beta_1}-\bar{E}_{\alpha_1,\beta_1}V'
    _{\alpha_2,\beta_2,\alpha_1,\beta_1'}.
    \end{aligned}
    \]
    \item Backward recursion (part 2):
    \[
    V'_{\alpha_2,\beta_2,\alpha_1,\beta_1}=\min_{p,q}\sbk{\prs{\tilde{E}_{\alpha_1,\beta_1,\alpha_2,\beta_2}-(pE_{\alpha_1,\beta_1}+qE_{\alpha_2,\beta_2})}^2-p(V'_{\alpha_1,\beta_1,\alpha_2,\beta_2}-V'_{\alpha_2,\beta_2,\alpha_1,\beta_1})}.
    \]
    \item Boundary condition when arm 1 is known ($\mathrm{Beta}(\alpha_1,\beta_1)$ becomes $\delta_{\alpha_1/(\alpha_1+\beta_1)}$ as $\alpha_1+\beta_1\gti$):
    \[
    V'_{\alpha_1,\beta_1,\alpha_2,\beta_2}=0.
    \]
    \item Boundary condition when arm 2 is known ($\mathrm{Beta}(\alpha_2,\beta_2)$ becomes $\delta_{\alpha_2/(\alpha_2+\beta_2)}$ as $\alpha_2+\beta_2\gti$):
    \[
    V'_{\alpha_1,\beta_1,\alpha_2,\beta_2}=\min_{p,q}\sbk{\frac{\prs{\tilde{E}_{\alpha_1,\beta_1,\alpha_2,\beta_2}-(pE_{\alpha_1,\beta_1}+qE_{\alpha_2,\beta_2})}^2}{p}}.
    \]
\end{itemize}
\end{proposition}
Given the backward recursion and the two boundary conditions, a natural approximation scheme for implementing the $\mathcal{R}^2$-optimal policy is as follows: (i) for $\bar{M}<\infty$, impose the two boundary conditions on $\{(\alpha_1,\beta_1,\alpha_2,\beta_2):\alpha_1+\beta_1=\bar{M}\;\text{or}\;\alpha_2+\beta_2=\bar{M}\},$
i.e., for $k=1,2$, replace $\mathrm{Beta}(\alpha_k,\beta_k)$ with $\delta_{\alpha_k/(\alpha_k+\beta_k)}$ whenever $\alpha_k+\beta_k$ reaches $\bar{M}$;
and (ii) propagate the values of $V'$ inward via the backward recursion (part 1 for the difference, part 2 for the two values).

\begin{figure}[ht]
    \centering
    \begin{minipage}{0.49\textwidth}
        \centering
        \includegraphics[width=0.9\linewidth]{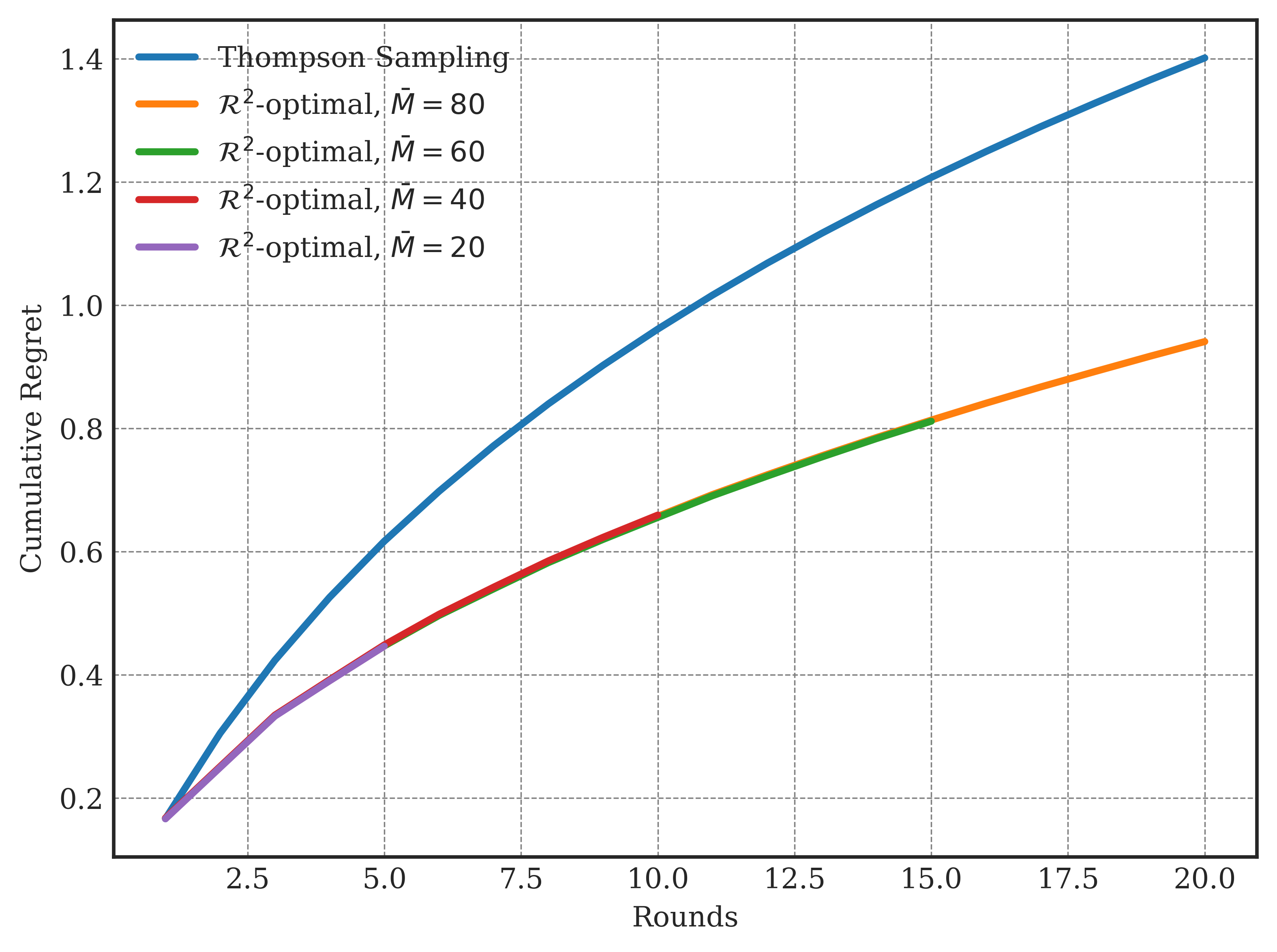}
    \end{minipage}
    \hfill
    \begin{minipage}{0.49\textwidth}
        \centering
        \includegraphics[width=0.9\linewidth]{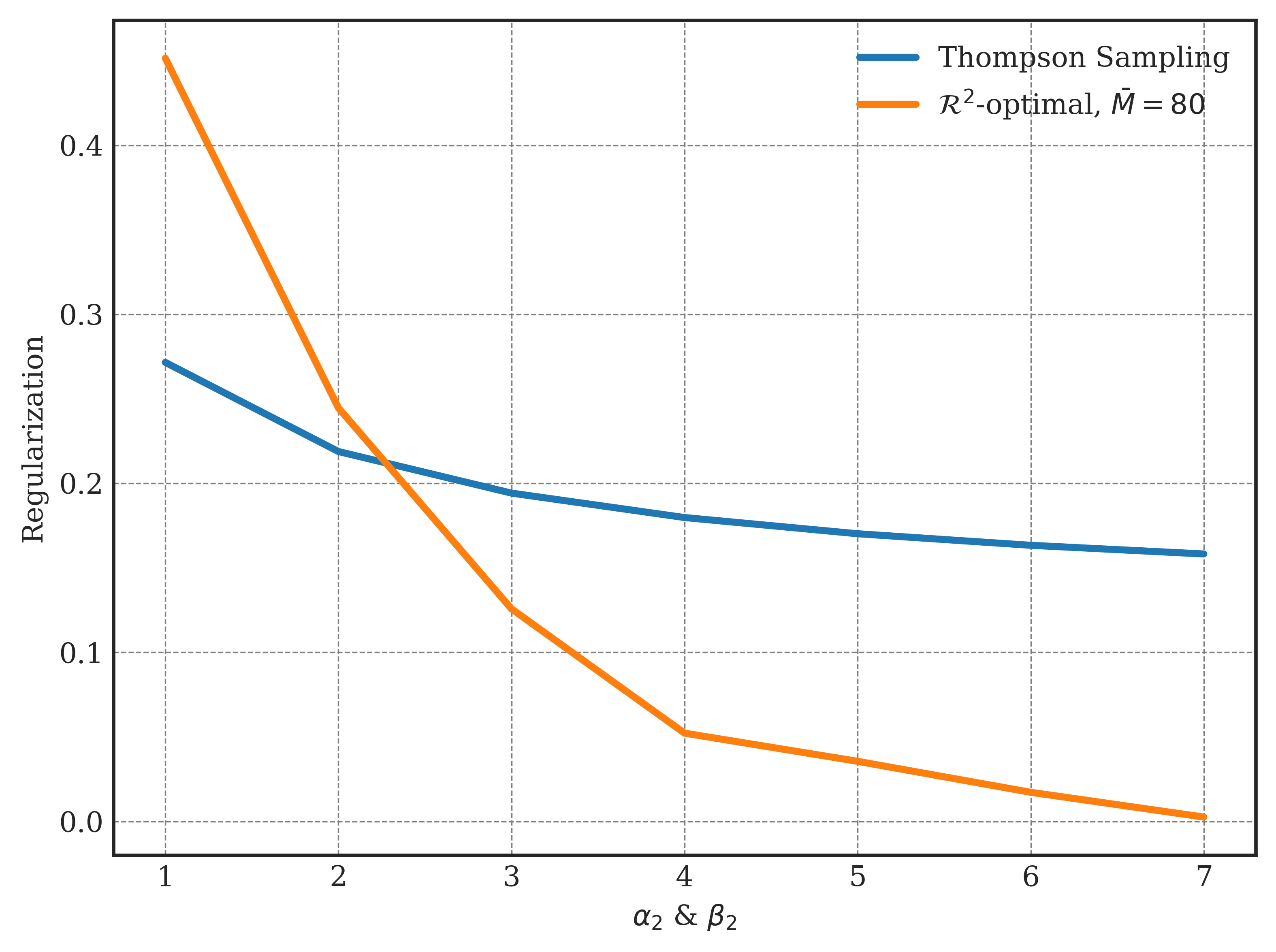}
    \end{minipage}

    \caption{
    Thompson Sampling and the $\mathcal{R}^2$-optimal policy (with different values of $\bar{M}$) play a Bernoulli bandit. Left: Comparing their cumulative regret $\mathcal{R}_T(Q^\text{TS};\pi_0)$ vs. $\mathcal{R}_T(Q^\text{R2};\pi_0)$ where $\pi_0=\mathrm{Beta}(1,1)\times\mathrm{Beta}(1,1)$ (200K trials). Right: Comparing their regularizers $\nu^\text{TS}(\mathrm{Beta}(5,4)\times\mathrm{Beta}(k,k))$ vs. $\nu^\text{R2}(\mathrm{Beta}(5,4)\times\mathrm{Beta}(k,k))$ where $k=1,\dots,7$.
    }
    \label{figure_6_ber}
\end{figure}

In Figure \ref{figure_6_ber}, we compare Thompson Sampling with the $\mathcal{R}^2$-optimal policy (with different values of $\bar{M}$) in the two-armed case. For each value of $\bar{M}$, let the corresponding policy play $\bar{M}/4$ rounds. The left panel of Figure \ref{figure_6_ber} shows that the regret curves corresponding to different values of $\bar{M}$ are closely aligned, indicating that the $\bar{M}$-truncation already captures the behavior of the $\mathcal{R}^2$-optimal policy over the first $\bar{M}/4$ rounds. After 20 rounds, the $\mathcal{R}^2$-optimal policy achieves a 30\% reduction in cumulative regret relative to Thompson Sampling. Again, the reason behind the gap can be understood via regularizer comparison. In the right panel of Figure \ref{figure_6_ber}, we compare the two regularizers along the prior sequence $\pi_0(k)=\mathrm{Beta}(5,4)\times\mathrm{Beta}(k,k)$, with $k=1,\dots,7$, to illustrate how the two policies respond as the under-performing arm gradually becomes over-explored. For arm 1, the mean of $\mathrm{Beta}(5,4)$ exceeds $1/2$. For arm 2, the mean of $\mathrm{Beta}(k,k)$ remains $1/2$ (under-performing), while the distribution becomes increasingly concentrated (over-explored) as $k$ increases. In particular, when $\mathrm{Beta}(5,4)$ vs. $\mathrm{Beta}(7,7)$, arm 1 receives fewer pulls ($5+4<7+7$) but has a higher posterior mean reward $5/9$, making it attractive from both exploration and exploitation perspectives. The right panel of Figure \ref{figure_6_ber} shows that the regularizer of the $\mathcal{R}^2$-optimal policy decays quickly, thereby encouraging fully greedy behavior (i.e., pulling arm 1 with probability one). In contrast, the regularizer of Thompson Sampling does not decay fast enough to temporarily stop pulling arm 2, which is under-performing and over-explored. Again, this conservativeness is because the uncertainty measure $\nu^\mathrm{TS}(\pi_t)$ does not fully capture the tension between exploration and exploitation quantified by $\nu^\mathrm{R2}(\pi_t)$. As the under-performing arm gradually becomes over-explored, the tension vanishes while the uncertainty (now primarily contributed by the other arm) remains.

\subsection{Information-Directed Sampling}
Recall that, through the online optimization lens, Thompson Sampling regularizes greediness according to uncertainty, whereas the $\mathcal{R}^2$-optimal policy does so according to tension.
The comparative analysis so far shows that the remaining uncertainty (about which arm is better) is {\it not} a sufficient proxy for the current tension (between exploration and exploitation), which in turn leads to a substantial performance gap between the two policies. What about Information-Directed Sampling (IDS)? Recall that IDS \eqref{eqn_ids} regularizes greediness according to tension \eqref{eqn_ids_reg}, suggesting that it may be competitive vis-a-vis the $\mathcal{R}^2$-optimal policy. (To avoid introducing additional notation, we adopt the variance-based information gain $\mathcal{I}(q_t;\pi_t)=q_t\cdot\var_{\pi_t}\E_{\pi_t}(\theta|\Lambda)$.)

\begin{figure}[ht]
    \centering
    \begin{minipage}{0.49\textwidth}
        \centering
        \includegraphics[width=0.9\linewidth]{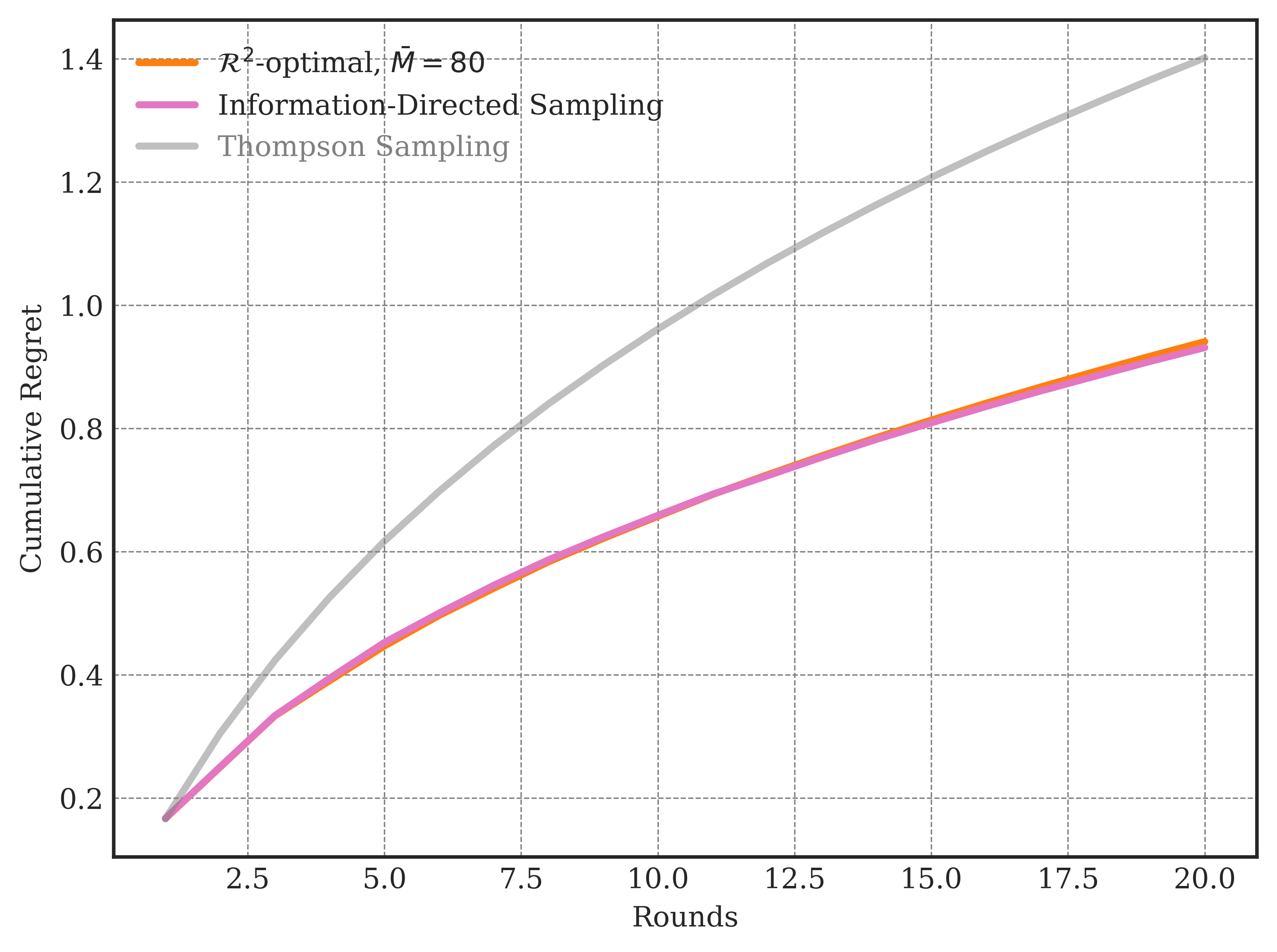}
    \end{minipage}
    \hfill
    \begin{minipage}{0.49\textwidth}
        \centering
        \includegraphics[width=0.9\linewidth]{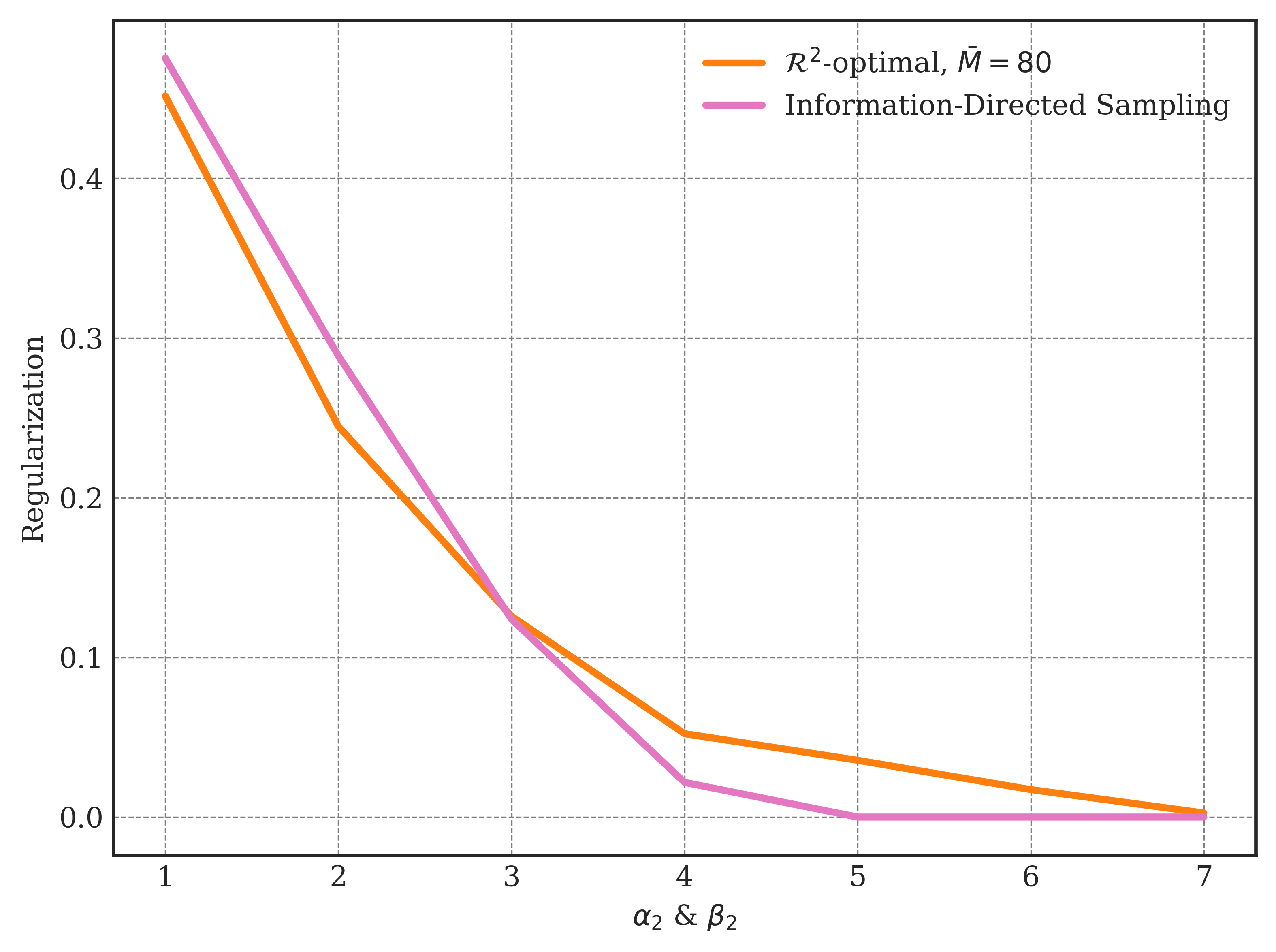}
    \end{minipage}

    \caption{
    Information-Directed Sampling and the $\mathcal{R}^2$-optimal policy (with $\bar{M}=80$) play a Bernoulli bandit. Left: Comparing their cumulative regret $\mathcal{R}_T(Q^\text{IDS};\pi_0)$ vs. $\mathcal{R}_T(Q^\text{R2};\pi_0)$ where $\pi_0=\mathrm{Beta}(1,1)\times\mathrm{Beta}(1,1)$ (200K trials). Right: Comparing their regularizers $\nu^\text{IDS}(\mathrm{Beta}(5,4)\times\mathrm{Beta}(k,k))$ vs. $\nu^\text{R2}(\mathrm{Beta}(5,4)\times\mathrm{Beta}(k,k))$ where $k=1,\dots,7$.
    }
    \label{figure_6_ids}
\end{figure}

In Figure \ref{figure_6_ids}, we compare IDS with the $\mathcal{R}^2$-optimal policy. The left panel shows that the two policies achieve nearly indistinguishable cumulative regret, and the right panel shows that their regularizers are highly aligned. This is surprising as the two policies originate from entirely different worlds: information-theoretic analysis vs. dynamic programming.
Recall that the goal of faithful stationarization is to derive a policy from first principles that achieves what Thompson Sampling achieves, thereby shedding light on the optimization considerations underlying this simple heuristic \citep{thompson1933likelihood}. The resulting principled policy turns out to go beyond Thompson Sampling in theory and empirically arrives at IDS, an advanced heuristic based on the information-ratio proof technique \citep{russo2016information}. Hence the optimization lens bridges two seminal ideas that emerged decades apart, revealing a broader view of Thompson Sampling.  In one case, information-theoretic analysis allowed a heuristic leap  from Thompson Sampling to IDS. In contrast, dynamic programming offers a principled path for improving Thompson Sampling: policy improvement.

\begin{remark}[So which policy?]
The $\mathcal{R}^2$-optimal policy and IDS appear to be tied over the first 20 rounds, beyond which approximation error (associated with $\bar{M}=80$) may begin to play a role. Since $\mathcal{R}^2(Q^\mathrm{R2};\pi_0)\leq\mathcal{R}^2(Q^\mathrm{IDS};\pi_0)$, $Q^\mathrm{R2}$ has better constant than $Q^\mathrm{IDS}$ in the regret bound \eqref{eqn_bound}, but this does not guarantee $\mathcal{R}_T(Q^\mathrm{R2};\pi_0)\leq\mathcal{R}_T(Q^\mathrm{IDS};\pi_0)$. Further analysis of this pair is left for future research, where the ``two-world clash'' may yield new insights.
\end{remark}

\section{Policy Improvement}
\label{sec_pi}
In reinforcement learning, the default objective is maximizing cumulative discounted reward, for which policy improvement is a standard way to obtain improved  policies. Once a policy has been evaluated, plugging its value function into the right-hand side
of the Bellman equation \eqref{eqn_Bellman_dis} produces a new policy that achieves higher cumulative discounted reward. In the bandit setting, however, the ultimate goal is long-term regret minimization, so it remains unclear how to perform policy improvement until our faithful stationarization yields the Bellman equation \eqref{eqn_Bellman}.

For Thompson Sampling, let $V^\mathrm{TS}(\cdot)=\mathcal{R}^2(Q^\mathrm{TS};\cdot)$ be its squared-regret value function, which can be defined for any $\mathcal{R}^2$-finite policy.
Plugging this value function into the right-hand side of the Bellman equation \eqref{eqn_Bellman} produces the one-step improved Thompson Sampling
\[
q_t^{\mathrm{TS}'}=\underset{q_t}{\mathrm{argmin}}\sbk{r^2(q_t;\pi_t)+q_t\cdot\E_{\pi_t}\sbk{V^\mathrm{TS}(\pi_{t+1})|A_t=\cdot}},
\]
which achieves lower squared regret than Thompson Sampling. This is because
\[
\begin{aligned}
V^{\mathrm{TS}}(\pi_t)
=&
r^2(q_t^{\mathrm{TS}};\pi_t)
+
q_t^{\mathrm{TS}}\cdot\E_{\pi_t}\sbk{V^{\mathrm{TS}}(\pi_{t+1})| A_t=\cdot} \\
\geq&
r^2(q_t^{\mathrm{TS}'};\pi_t)
+
q_t^{\mathrm{TS}'}\cdot\E_{\pi_t}\sbk{V^{\mathrm{TS}}(\pi_{t+1})| A_t=\cdot}\\
\geq&
r^2(q_t^{\mathrm{TS}'};\pi_t)
+
q_t^{\mathrm{TS}'}\cdot\E_{\pi_t}\sbk{r^2(q_{t+1}^{\mathrm{TS}'};\pi_{t+1})| A_t=\cdot}\\
&+q_t^{\mathrm{TS}'}\cdot\E_{\pi_t}\sbk{
q_{t+1}^{\mathrm{TS}'}\cdot\E_{\pi_{t+1}}\sbk{V^{\mathrm{TS}}(\pi_{t+2})| A_{t+1}=\cdot}|A_t=\cdot}\\
\geq&...\geq V^{\mathrm{TS}'}(\pi_t),
\end{aligned}
\]
where the first equality is the Bellman equation for policy evaluation, the first inequality follows from the definition of $q_t^{\mathrm{TS}'}$, and the rest unfolds through iteration.
If we apply another policy-improvement step to $Q^{\mathrm{TS}'}$, we obtain $Q^{\mathrm{TS}''}$, and continuing in this way (i.e., policy iteration) yields
\[
\mathcal{R}^2(Q^\mathrm{TS};\cdot)\geq\mathcal{R}^2(Q^{\mathrm{TS}'};\cdot)\geq\mathcal{R}^2(Q^{\mathrm{TS}''};\cdot)\geq...\geq\mathcal{R}^2(Q^\mathrm{R2};\cdot),
\]
which successively reduces the leading constant in the regret bound \eqref{eqn_bound}. A natural question is: how many steps are needed to make Thompson Sampling achieve performance that is comparable to the Bellman-optimal benchmark? Remarkably, a single step is essentially sufficient. 

\begin{figure}[ht]
    \centering
    \begin{minipage}{0.49\textwidth}
        \centering
        \includegraphics[width=0.9\linewidth]{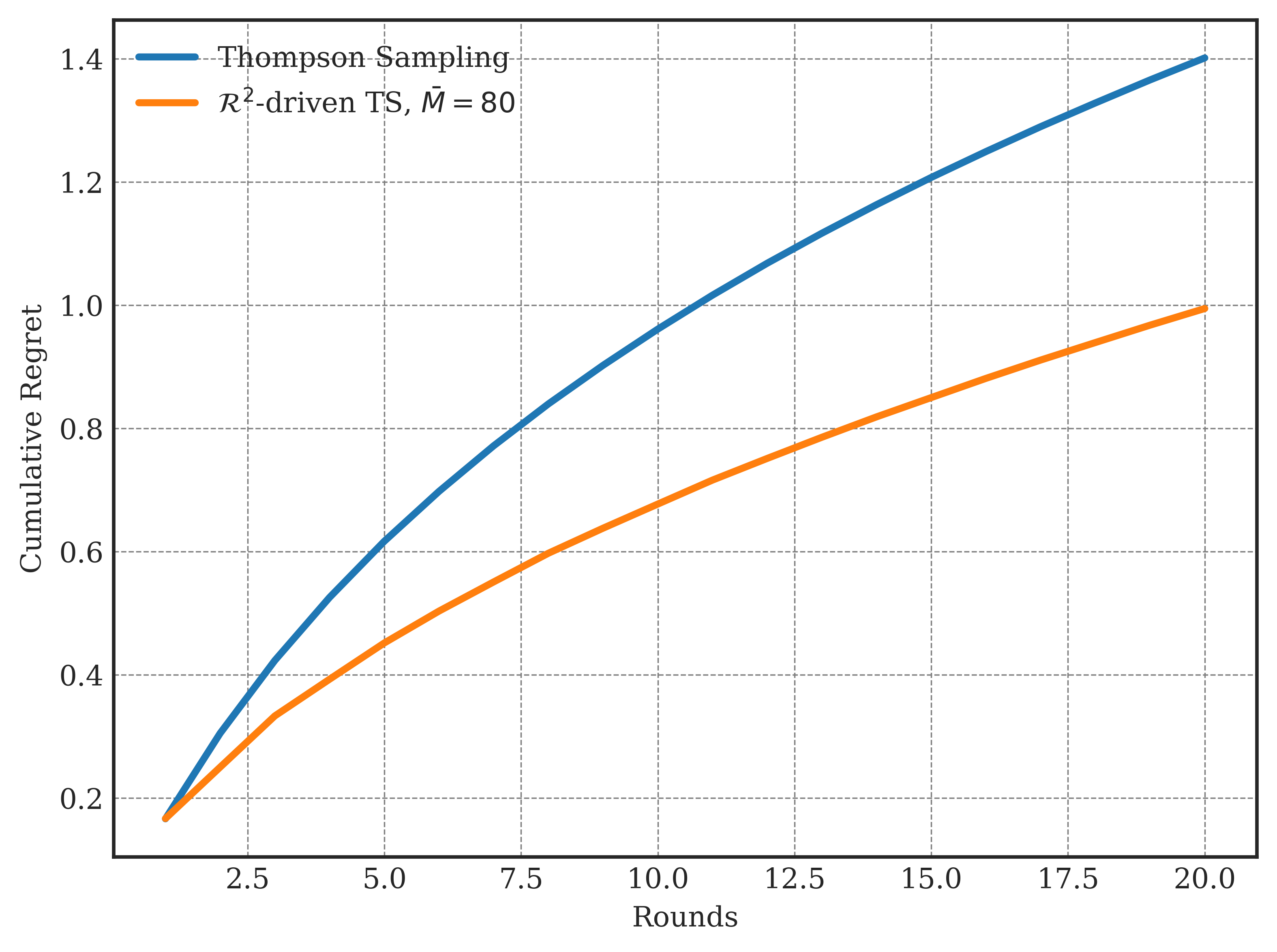}
    \end{minipage}
    \hfill
    \begin{minipage}{0.49\textwidth}
        \centering
        \includegraphics[width=0.9\linewidth]{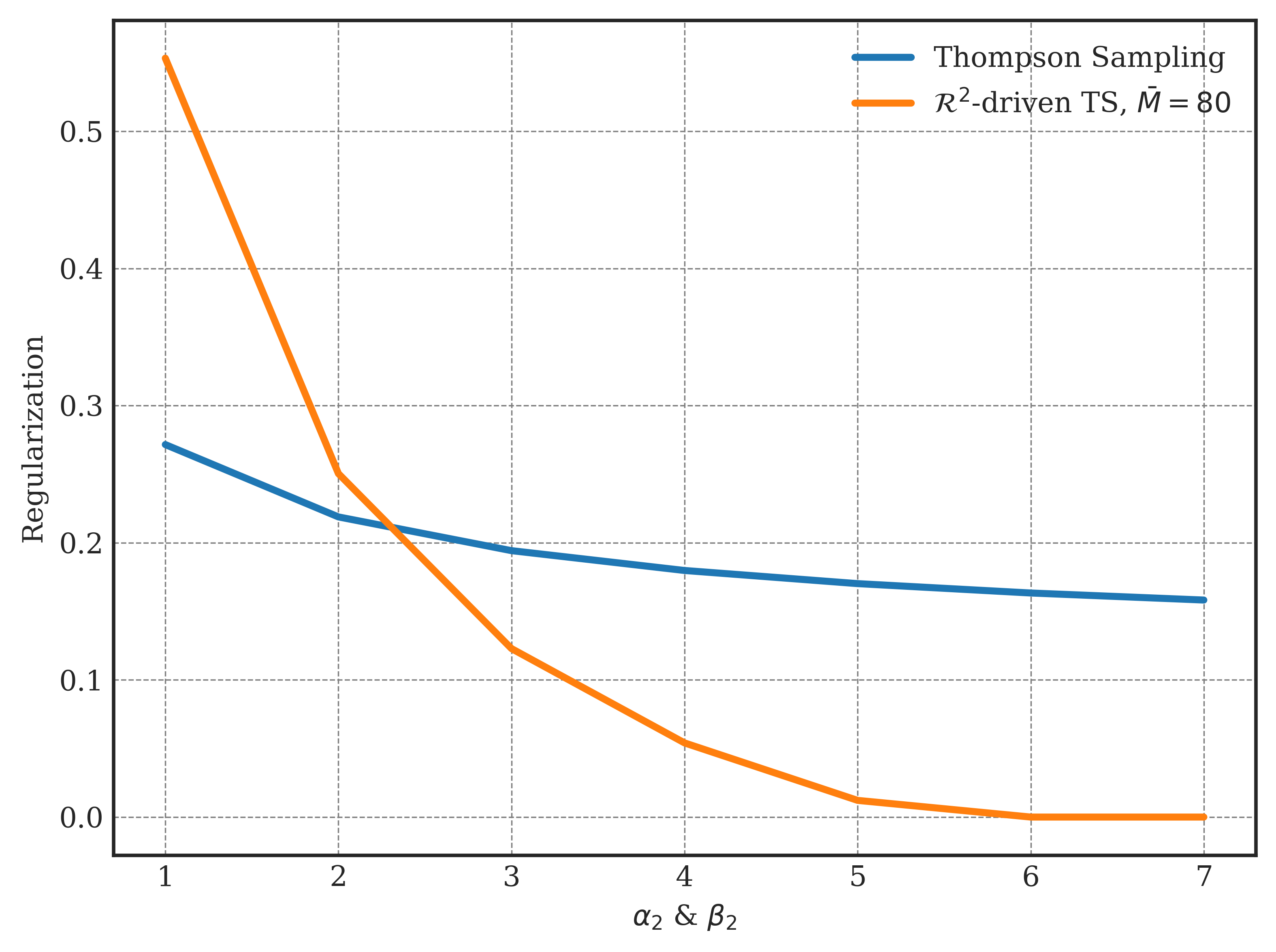}
    \end{minipage}

    \caption{
    Thompson Sampling and its one-step improved counterpart (with $\bar{M}=80$) play a Bernoulli bandit. Left: Comparing their cumulative regret $\mathcal{R}_T(Q^\text{TS};\pi_0)$ vs. $\mathcal{R}_T(Q^{\text{TS}'};\pi_0)$ where $\pi_0=\mathrm{Beta}(1,1)\times\mathrm{Beta}(1,1)$ (200K trials). Right: Comparing their regularizers $\nu^\text{TS}(\mathrm{Beta}(5,4)\times\mathrm{Beta}(k,k))$ vs. $\nu^{\text{TS}'}(\mathrm{Beta}(5,4)\times\mathrm{Beta}(k,k))$ where $k=1,\dots,7$.
    }
    \label{figure_7_ber}
\end{figure}

In Figure \ref{figure_7_ber}, we compare Thompson Sampling pre and post a single policy-improvement step. The striking similarity between Figure \ref{figure_7_ber} ($Q^\mathrm{TS}$ vs. $Q^{\mathrm{TS}'}$) and Figure \ref{figure_6_ber} ($Q^\mathrm{TS}$ vs. $Q^\mathrm{R2}$) shows that this single step already brings Thompson Sampling quite close to the Bellman-optimal benchmark. The left panel of Figure \ref{figure_7_ber} shows that this single step closes a large proportion (about 90\%) of the performance gap between Thompson Sampling and the Bellman-optimal benchmark. The right panel of Figure \ref{figure_7_ber} shows that this single step essentially transforms the regularizer of Thompson Sampling from an uncertainty measure to a tension measure
\[
\nu^{\mathrm{TS}'}(\pi_t)=\sbk{\frac{\E_{\pi_t}\sbk{V^\mathrm{TS}(\pi_{t+1})|A_t=1}-\E_{\pi_t}\sbk{V^\mathrm{TS}(\pi_{t+1})|A_t=2}}{\E_{\pi_t}\theta_1-\E_{\pi_t}\theta_2}}_+,
\]
which explains much of the performance gain.
Altogether, this example illustrates the power of policy improvement, guided by first principles.

\begin{remark}[Implementing policy evaluation]
    For two-armed Bernoulli bandits, we can borrow the $\bar{M}$-truncation from Section \ref{sec_tvb} to approximately perform policy evaluation. More generally, multi-armed bandit policy evaluation, improvement, and iteration at scale (e.g., storing  $V^\mathrm{TS}$, and hence $Q^{\mathrm{TS}'}$, in a neural network) is a natural direction for future research.
\end{remark}

\subsection{More than two arms}
Why does Thompson Sampling perform so well after a single policy-improvement step, landing in the neighborhood of the Bellman-optimal benchmark? The change here must be structural rather than incremental. In the two-armed case, it is the exploration logic (i.e., the regularization mechanism) of Thompson Sampling that is upgraded from uncertainty-driven to tension-driven. When there are more than two arms, the structural change turns out to be even more fundamental.

In the $K$-armed case ($K>2$), Thompson Sampling pulls an arm according to its probability of being optimal, so every arm receives positive pulling probability unless it is known to be suboptimal with certainty. In stark contrast, any policy of the online optimization form \eqref{eqn_q_form_r2}, including not only $Q^\mathrm{R2}$ but also $Q^{\mathrm{TS}'}$ (and even $Q^\mathrm{IDS}$), assigns positive pulling probability to at most two arms at each round. To see this, let us take $Q^\mathrm{R2}$ as an example. Recall that $x_t=q_t\cdot\E_{\pi_t}\theta$ is the expected next-round reward. After the change of variables, the online optimization form \eqref{eqn_q_form_r2} becomes
\[
x_t^\mathrm{R2}=\underset{x_t}{\mathrm{argmin}}\sbk{\prs{\E_{\pi_t}\max\{\theta_1,...\theta_K\}-x_t}^2+h(x_t;\pi_t)},
\]
where
\[
h(x_t;\pi_t)=\min\left\{q_t\cdot\E_{\pi_t}\sbk{V(\pi_{t+1})|A_t=\cdot}:\;q_t\geq0,\;q_t\cdot\mathbf{1}=1,\;q_t\cdot\E_{\pi_t}\theta=x_t\right\}.
\]
Given $\pi_t$, $h(x_t;\pi_t)$ is the lower convex envelope of the following $K$ points
\[
\left\{(\E_{\pi_t}\theta_k,\E_{\pi_t}\sbk{V(\pi_{t+1})|A_t=k}):k=1,...,K\right\},
\]
which is a piecewise linear function of $x_t$. Given $x_t$, the point $(x_t,h(x_t;\pi_t))$ lies on either an edge or a vertex of the envelope, so the corresponding $q_t$ mixes at most two vertices (i.e., arms). 

\begin{figure}[ht]
    \centering
    \begin{minipage}{0.49\textwidth}
        \centering
        \includegraphics[width=0.9\linewidth]{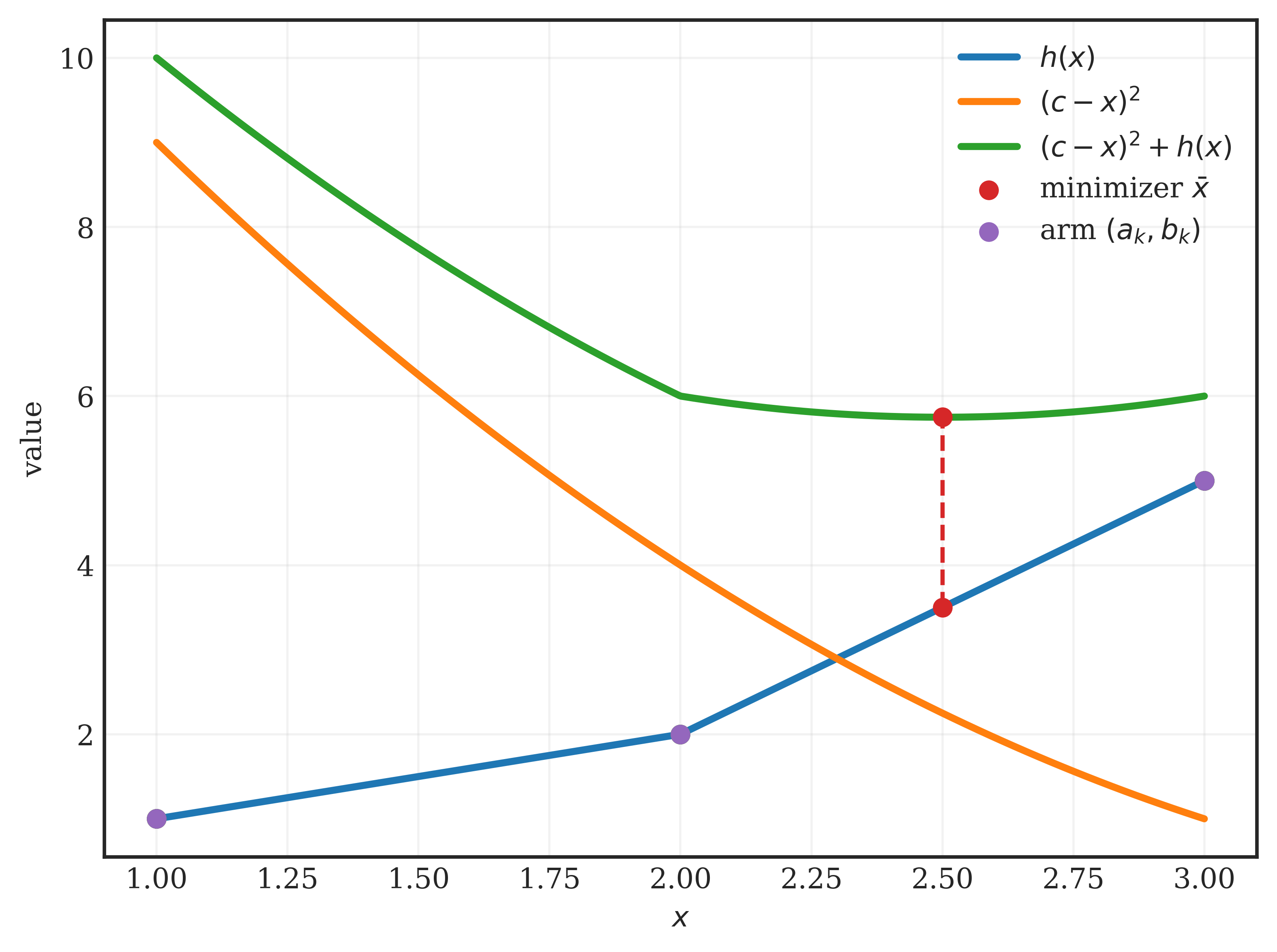}
    \end{minipage}
    \hfill
    \begin{minipage}{0.49\textwidth}
        \centering
        \includegraphics[width=0.9\linewidth]{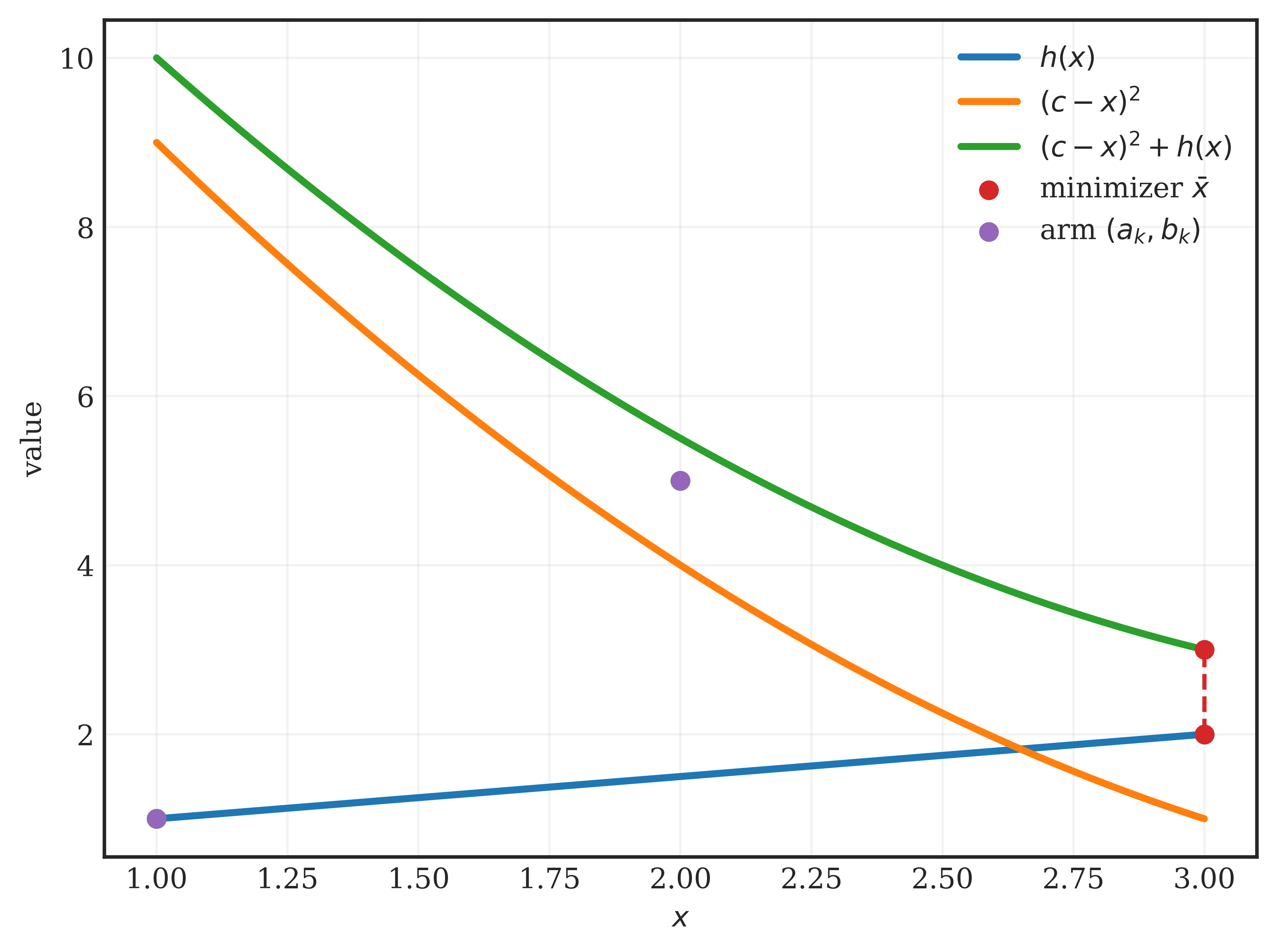}
    \end{minipage}

    \caption{Two examples of recovering the minimizer of $(c-a\cdot q)^2+b\cdot q$ from the minimizer of $(c-x)^2+h(x)$ where $h$ is the lower convex envelope. Left: $a=(1,2,3)$, $b=(1,2,5)$, $c=4$. Right: $a=(1,2,3)$, $b=(1,5,2)$, $c=4$.
    }
    \label{figure_7_two}
\end{figure}

In Figure \ref{figure_7_two}, we visualize the geometry of minimizing
\[
(c-a\cdot q)^2+b\cdot q\quad\text{or}\quad(c-x)^2+h(x),
\]
where $h$ is the lower convex envelope of $\{(a_1,b_1),(a_2,b_2),(a_3,b_3)\}$ (from left to right in the plots). In the left panel of Figure \ref{figure_7_two}, the minimizer $\bar{x}$ of $(c-x)^2+h(x)$ is between $a_2$ and $a_3$, so the point $(\bar{x},h(\bar{x}))$ is on the edge joining $(a_2,b_2)$ and $(a_3,b_3)$. For this $\bar{x}$, the corresponding $\bar{q}$ is $(0,1/2,1/2)$. In the right panel of Figure \ref{figure_7_two}, the minimizer $\bar{x}$ of $(c-x)^2+h(x)$ is exactly $a_3$, so the point $(\bar{x},h(\bar{x}))$ is on the vertex $(a_3,b_3)$. For this $\bar{x}$, the corresponding $\bar{q}$ is $(0,0,1)$.

When viewed through the lens of online optimization, we indeed obtain a {\it visualization} of the MAB problem (Figure \ref{figure_7_two}), where everything is compiled into a single two-dimensional plot, regardless of the number of arms. Moreover, the key decision variable, the expected next-round reward, is a scalar, from which one can recover the probability vector assigned to the arms, which contains at most two positive entries. Effectively, in our squared regret formulation, Bellman's principle suggests that mixing at most two arms is a highly desirable structural property, and policy improvement grants Thompson Sampling this property in a single step. 

\section{Proofs}
\label{sec_proof}

\begin{proof}{Proof of Proposition \ref{prop_ts_finite}.}
By the information-ratio bound of \cite{russo2016information}, under the stated $\sigma$-sub-Gaussian assumption, we have
\[
r^2(q_t^{\mathrm{TS}};\pi_t)\le 2K\sigma^2 \cdot\E_{\pi_t}\sbk{D_\mathrm{KL}(q_{t+1}^\mathrm{TS}\|q_t^\mathrm{TS})},
\]
where $K$ is the number of arms, and $D_\mathrm{KL}(\cdot\|\cdot)$ is the KL divergence.
Since
\[
\E_{\pi_t}q_{t+1}^\mathrm{TS}=\E_{\pi_t}P_{\pi_t}\prs{\mathrm{argmax}\{\theta_1,...\theta_K\}=\cdot|\pi_{t+1}}=P_{\pi_t}\prs{\mathrm{argmax}\{\theta_1,...\theta_K\}=\cdot}=q_t^\mathrm{TS},
\]
we have
\[
\begin{aligned}
    \E_{\pi_t}\sbk{D_\mathrm{KL}(q_{t+1}^\mathrm{TS}\|q_t^\mathrm{TS})}=&\E_{\pi_t}\sbk{\sum_{k=1}^K q_{k,t+1}^\mathrm{TS}\log\frac{q_{k,t+1}^\mathrm{TS}}{q_{k,t}^\mathrm{TS}}}\\
=&\E_{\pi_t}\sbk{\sum_{k=1}^K q_{k,t+1}^\mathrm{TS}\log q_{k,t+1}^\mathrm{TS}}-\sum_{k=1}^K q_{k,t}^\mathrm{TS}\log q_{k,t}^\mathrm{TS}\\
=&H(q_t^\mathrm{TS})-\E_{\pi_t}H(q_{t+1}^\mathrm{TS}),
\end{aligned}
\]
where $H(\cdot)$ is the entropy. Therefore, we have
\[
\begin{aligned}
\mathcal{R}^2(Q^\mathrm{TS};\pi_0)=&\E_{\pi_0}\sbk{\sum_{t=0}^{\infty}r^2(q_t^\mathrm{TS};\pi_t)}\\
\leq& 2K\sigma^2\cdot\E_{\pi_0}\sbk{\sum_{t=0}^\infty \sbk{H(q_t^\mathrm{TS})-\E_{\pi_t}H(q_{t+1}^\mathrm{TS})}}\\
=&2K\sigma^2\cdot \sum_{t=0}^\infty \sbk{\E_{\pi_0}H(q_t^\mathrm{TS})-\E_{\pi_0}H(q_{t+1}^\mathrm{TS})}\\
\leq&2K\sigma^2\cdot H(q_0^\mathrm{TS})\\
<&\infty.
\end{aligned}
\]
\end{proof}

\begin{proof}{Proof of Theorem \ref{thm_ts}.}
Recall that $\Delta=\theta_1-\theta_2$, $\Lambda=\mathrm{sign}(\theta_1-\theta_2)$, and
\[
\begin{aligned}
    \frac{\cov_{\pi_t}(\Delta,\Lambda)}{2}=&\E_{\pi_t}\Delta I(\Delta>0)-P_{\pi_t}(\Delta>0)\E_{\pi_t}\Delta\\
    =&P_{\pi_t}(\Delta\leq0)\E_{\pi_t}\Delta I(\Delta>0)+P_{\pi_t}(\Delta>0)\E_{\pi_t}\Delta I(\Delta>0)\\
    &-P_{\pi_t}(\Delta>0)\E_{\pi_t}\Delta I(\Delta>0)-P_{\pi_t}(\Delta>0)\E_{\pi_t}\Delta I(\Delta\leq0)\\
    =&P_{\pi_t}(\Delta\leq0)\E_{\pi_t}\Delta I(\Delta>0)-P_{\pi_t}(\Delta>0)\E_{\pi_t}\Delta I(\Delta\leq0).
\end{aligned}
\]
By differentiation, the minimizer of the online objective is
\[
\begin{aligned}
    &\E_{\pi_t}\max(\theta_1,\theta_2)-\frac{\cov_{\pi_t}(\Delta,\Lambda)}{2}\\
    =&P_{\pi_t}(\Delta\leq0)(\E_{\pi_t}\theta_2+\E_{\pi_t}\Delta I(\Delta>0))+P_{\pi_t}(\Delta>0)(\E_{\pi_t}\theta_1-\E_{\pi_t}\Delta I(\Delta\leq0))\\
    &-P_{\pi_t}(\Delta\leq0)\E_{\pi_t}\Delta I(\Delta>0)+P_{\pi_t}(\Delta>0)\E_{\pi_t}\Delta I(\Delta\leq0)\\
    =&P_{\pi_t}(\Delta>0)\E_{\pi_t}\theta_1+P_{\pi_t}(\Delta\leq0)\E_{\pi_t}\theta_2,
\end{aligned}    
\]
which is the expected next-round reward of Thompson Sampling.
\end{proof}
\vspace{0.5em}
\begin{proof}{Proof of Proposition \ref{prop_incomplete}.}
For $\lambda\neq1$, the unconstrained minimizer of the online objective is
\[
\begin{aligned}
    \bar{x}_t^\lambda=&\E_{\pi_t}\max(\theta_1,\theta_2)-\frac{\lambda\cov_{\pi_t}(\Delta,\Lambda)}{2}\\
    =&\lambda\prs{\E_{\pi_t}\max(\theta_1,\theta_2)-\frac{\cov_{\pi_t}(\Delta,\Lambda)}{2}}+(1-\lambda)\E_{\pi_t}\max(\theta_1,\theta_2)\\
    =&\lambda\prs{P_{\pi_t}(\Delta>0)\E_{\pi_t}\theta_1+P_{\pi_t}(\Delta\leq0)\E_{\pi_t}\theta_2}+(1-\lambda)\E_{\pi_t}\max(\theta_1,\theta_2).
\end{aligned}
\]
The constrained minimizer $x_t^\lambda$ is obtained by clipping $\bar{x}_t^\lambda$ to be between $\E_{\pi_t}\theta_1$ and $\E_{\pi_t}\theta_2$. When $\pi_t= \delta_{1-\lambda}\times N(0,\sigma^2)$, we have
\[
    \E_{\pi_t}\max(\theta_1,\theta_2)=\E \max(1-\lambda,N(0,\sigma^2))=\sigma\E\max((1-\lambda)/\sigma,N(0,1))\gti,
\]
as $\sigma\gti$. When $\sigma$ is large enough, we have
\[
\begin{aligned}
    &\lambda<1\Rightarrow \bar{x}_t^\lambda>\E_{\pi_t}\theta_1>\E_{\pi_t}\theta_2\Rightarrow x_t^\lambda=\E_{\pi_t}\theta_1,\\
    &\lambda>1\Rightarrow \bar{x}_t^\lambda<\E_{\pi_t}\theta_1<\E_{\pi_t}\theta_2\Rightarrow x_t^\lambda=\E_{\pi_t}\theta_1.
\end{aligned}
\]
In either case, arm 1 is pulled with probability one, but pulling the known arm 1 produces no posterior update. Consequently, the policy fully commits to arm 1 while arm 2 still has a chance of being better. 
\end{proof}
\vspace{0.5em}
\begin{proof}{Proof of Proposition \ref{prop_cov}.}
When $\var_{\pi_t}\Lambda>0$, we have
\[
\begin{aligned}
    \frac{\cov_{\pi_t}(\Delta,\Lambda)}{\var_{\pi_t}\Lambda}=&\frac{2P_{\pi_t}(\Delta\leq0)\E_{\pi_t}\Delta I(\Delta>0)}{4 P_{\pi_t}(\Delta>0)P_{\pi_t}(\Delta\leq0)}-\frac{2P_{\pi_t}(\Delta>0)\E_{\pi_t}\Delta I(\Delta\leq0)}{4 P_{\pi_t}(\Delta>0)P_{\pi_t}(\Delta\leq0)}\\
    =&\frac{\E_{\pi_t}[\Delta|\Delta>0]+\E_{\pi_t}[-\Delta|\Delta\leq0]}{2}.
\end{aligned}
\]    
\end{proof}
\vspace{0.5em}
\begin{proof}{Proof of Proposition \ref{prop_one_arm}.}
The minimizer of 
\[
\frac{\prs{\E_{\pi_t}[(\theta_1)_+]-q_{1,t}\E_{\pi_t}\theta_1}^2}{q_{1,t}}=\frac{\prs{\E_{\pi_t}[(\theta_1)_+]}^2}{q_{1,t}}+q_{1,t}\prs{\E_{\pi_t}\theta_1}^2-2\E_{\pi_t}[(\theta_1)_+]\E_{\pi_t}\theta_1
\]
in $[0,1]$ is clearly
\[
q^\mathrm{R2}_{1,t}=\min\prs{\frac{\E_{\pi_t}[(\theta_1)_+]}{|\E_{\pi_t}\theta_1|},1}.
\]
\end{proof}
\vspace{0.5em}
\begin{proof}{Proof of Proposition \ref{prop_gaussian}.}
When $\theta_2\equiv0$ and $\theta_1\sim N(\mu_t,\sigma_t^2)$ under $\pi_t$, we have
\[
\begin{aligned}
    q^\mathrm{R2}_{1,t}=1\;\;\Leftrightarrow\;\;&\E_{\pi_t}[(\theta_1)_+]\geq|\E_{\pi_t}\theta_1|\\
\;\;\Leftrightarrow\;\;&\E_{\pi_t}[(\theta_1)_+]\geq-\E_{\pi_t}\theta_1\\
\;\;\Leftrightarrow\;\;&\E_{\pi_t}[(\theta_1)_+]+\E_{\pi_t}\theta_1\geq0\\
\;\;\Leftrightarrow\;\;&\mu_t\Phi\prs{\frac{\mu_t}{\sigma_t}}+\sigma_t\phi\prs{\frac{\mu_t}{\sigma_t}}+\mu_t\geq0\\
\;\;\Leftrightarrow\;\;&\frac{\mu_t}{\sigma_t}\Phi\prs{\frac{\mu_t}{\sigma_t}}+\phi\prs{\frac{\mu_t}{\sigma_t}}+\frac{\mu_t}{\sigma_t}\geq0\\
\;\;\Leftrightarrow\;\;&\frac{\mu_t}{\sigma_t}\geq\bar{x},
\end{aligned}
\]
where $\bar{x}\approx-0.276$ is the unique root of the increasing function $x\Phi(x)+\phi(x)+x$.
\end{proof}
\vspace{0.5em}
\begin{proof}{Proof of Proposition \ref{prop_benefit}.}
\textbf{Backward recursion (part 1).}
Note that both sides of
    \[
    \begin{aligned}
    &V'_{\alpha_1,\beta_1,\alpha_2,\beta_2}-E_{\alpha_2,\beta_2}V'_{\alpha_1,\beta_1,\alpha_2',\beta_2}-\bar{E}_{\alpha_2,\beta_2}V'
    _{\alpha_1,\beta_1,\alpha_2,\beta_2'}\\
    =&V'_{\alpha_2,\beta_2,\alpha_1,\beta_1}-E_{\alpha_1,\beta_1}V'_{\alpha_2,\beta_2,\alpha_1',\beta_1}-\bar{E}_{\alpha_1,\beta_1}V'
    _{\alpha_2,\beta_2,\alpha_1,\beta_1'},
    \end{aligned}
    \]
equal to
\[
\begin{aligned}
    &V_{\alpha_1,\beta_1,\alpha_2,\beta_2}-E_{\alpha_1,\beta_1}V_{\alpha_1',\beta_1,\alpha_2,\beta_2}-\bar{E}_{\alpha_1,\beta_1}V_{\alpha_1,\beta_1',\alpha_2,\beta_2}\\
&-E_{\alpha_2,\beta_2}V_{\alpha_1,\beta_1,\alpha_2',\beta_2}-\bar{E}_{\alpha_2,\beta_2}V
_{\alpha_1,\beta_1,\alpha_2,\beta_2'}\\
&+E_{\alpha_1,\beta_1}E_{\alpha_2,\beta_2}V_{\alpha_1',\beta_1,\alpha_2',\beta_2}+\bar{E}_{\alpha_1,\beta_1}\bar{E}_{\alpha_2,\beta_2}V_{\alpha_1,\beta_1',\alpha_2,\beta_2'}\\
&+\bar{E}_{\alpha_1,\beta_1}E_{\alpha_2,\beta_2}V_{\alpha_1,\beta_1',\alpha_2',\beta_2}+E_{\alpha_1,\beta_1}\bar{E}_{\alpha_2,\beta_2}V_{\alpha_1',\beta_1,\alpha_2,\beta_2'},
\end{aligned}
\]
which remains unchanged when subscripts 1 and 2 are swapped.

\textbf{Backward recursion (part 2).}
The Bellman equation for $V'$
\[
V'_{\alpha_2,\beta_2,\alpha_1,\beta_1}=\min_{p,q}\sbk{\prs{\tilde{E}_{\alpha_1,\beta_1,\alpha_2,\beta_2}-(pE_{\alpha_1,\beta_1}+qE_{\alpha_2,\beta_2})}^2-p(V'_{\alpha_1,\beta_1,\alpha_2,\beta_2}-V'_{\alpha_2,\beta_2,\alpha_1,\beta_1})}
\]
is obtained by subtracting $E_{\alpha_2,\beta_2}V_{\alpha_1,\beta_1,\alpha_2',\beta_2}+\bar{E}_{\alpha_2,\beta_2}V_{\alpha_1,\beta_1,\alpha_2,\beta_2'}$ from both sides of
\[
\begin{aligned}
    V_{\alpha_1,\beta_1,\alpha_2,\beta_2}=&\min_{p,q}\Big[\prs{\tilde{E}_{\alpha_1,\beta_1,\alpha_2,\beta_2}-(pE_{\alpha_1,\beta_1}+qE_{\alpha_2,\beta_2})}^2\\
    &\qquad\;\;+p(E_{\alpha_1,\beta_1}V_{\alpha_1',\beta_1,\alpha_2,\beta_2}+\bar{E}_{\alpha_1,\beta_1}V_{\alpha_1,\beta_1',\alpha_2,\beta_2})\\
    &\qquad\;\;+q(E_{\alpha_2,\beta_2}V_{\alpha_1,\beta_1,\alpha_2',\beta_2}+\bar{E}_{\alpha_2,\beta_2}V_{\alpha_1,\beta_1,\alpha_2,\beta_2'})\Big].
\end{aligned}
\]

\textbf{Boundary condition when arm 1 is known.} As pulling the known arm 1 no longer changes $V$, 
\[
V_{\alpha_1,\beta_1,\alpha_2,\beta_2}=V_{\alpha_1',\beta_1,\alpha_2,\beta_2}=V_{\alpha_1,\beta_1',\alpha_2,\beta_2}\;\;\Rightarrow\;\;V'_{\alpha_1,\beta_1,\alpha_2,\beta_2}=0.
\]

\textbf{Boundary condition when arm 2 is known.} When $V'_{\alpha_2,\beta_2,\alpha_1,\beta_1}=0$, the Bellman equation for $V'$ is equivalent to
\[
V'_{\alpha_1,\beta_1,\alpha_2,\beta_2}=\min_{p,q}\sbk{\frac{\prs{\tilde{E}_{\alpha_1,\beta_1,\alpha_2,\beta_2}-(pE_{\alpha_1,\beta_1}+qE_{\alpha_2,\beta_2})}^2}{p}}.
\]
\end{proof}
%
%
%






\bibliographystyle{informs2014} 
\bibliography{references} 

@article{thompson1933likelihood,
  title={On the likelihood that one unknown probability exceeds another in view of the evidence of two samples},
  author={Thompson, William R},
  journal={Biometrika},
  volume={25},
  number={3/4},
  pages={285--294},
  year={1933},
  publisher={JSTOR}
}

@article{chapelle2011empirical,
  title={An empirical evaluation of {T}hompson sampling},
  author={Chapelle, Olivier and Li, Lihong},
  journal={Advances in Neural Information Processing Systems},
  volume={24},
  year={2011}
}

@inproceedings{agrawal2012analysis,
  title={Analysis of {T}hompson sampling for the multi-armed bandit problem},
  author={Agrawal, Shipra and Goyal, Navin},
  booktitle={Conference on Learning Theory},
  pages={39--1},
  year={2012},
  organization={JMLR Workshop and Conference Proceedings}
}

@inproceedings{agrawal2013further,
  title={Further Optimal Regret Bounds for {T}hompson Sampling},
  author={Agrawal, Shipra and Goyal, Navin},
  booktitle={Artificial Intelligence and Statistics},
  pages={99--107},
  year={2013},
  organization={PMLR}
}

@article{russo2014learningIDS,
  title={Learning to optimize via information-directed sampling},
  author={Russo, Daniel and Van Roy, Benjamin},
  journal={Advances in Neural Information Processing Systems},
  volume={27},
  year={2014}
}

@article{russo2016information,
  title={An information-theoretic analysis of {T}hompson sampling},
  author={Russo, Daniel and Van Roy, Benjamin},
  journal={Journal of Machine Learning Research},
  volume={17},
  number={68},
  pages={1--30},
  year={2016}
}

@article{auer2002finite,
  title={Finite-time analysis of the multiarmed bandit problem},
  author={Auer, Peter and Cesa-Bianchi, Nicolo and Fischer, Paul},
  journal={Machine learning},
  volume={47},
  pages={235--256},
  year={2002},
  publisher={Springer}
}

@article{lai1985asymptotically,
  title={Asymptotically efficient adaptive allocation rules},
  author={Lai, Tze Leung and Robbins, Herbert},
  journal={Advances in Applied Mathematics},
  volume={6},
  number={1},
  pages={4--22},
  year={1985},
  publisher={Academic Press}
}

@article{scott2010modern,
  title={A modern {B}ayesian look at the multi-armed bandit},
  author={Scott, Steven L},
  journal={Applied Stochastic Models in Business and Industry},
  volume={26},
  number={6},
  pages={639--658},
  year={2010},
  publisher={Wiley Online Library}
}

@inproceedings{agarwal2013computational,
  title={Computational advertising: the {LinkedIn} way},
  author={Agarwal, Deepak},
  booktitle={Proceedings of the 22nd ACM International Conference on Information \& Knowledge Management},
  pages={1585--1586},
  year={2013}
}

@inproceedings{hill2017efficient,
  title={An efficient bandit algorithm for realtime multivariate optimization},
  author={Hill, Daniel N and Nassif, Houssam and Liu, Yi and Iyer, Anand and Vishwanathan, SVN},
  booktitle={Proceedings of the 23rd ACM SIGKDD International Conference on Knowledge Discovery and Data Mining},
  pages={1813--1821},
  year={2017}
}

@article{kawale2015efficient,
  title={Efficient {T}hompson Sampling for Online Matrix-Factorization Recommendation},
  author={Kawale, Jaya and Bui, Hung H and Kveton, Branislav and Tran-Thanh, Long and Chawla, Sanjay},
  journal={Advances in Neural Information Processing Systems},
  volume={28},
  year={2015}
}

@article{russo2014learningPS,
  title={Learning to optimize via posterior sampling},
  author={Russo, Daniel and Van Roy, Benjamin},
  journal={Mathematics of Operations Research},
  volume={39},
  number={4},
  pages={1221--1243},
  year={2014},
  publisher={INFORMS}
}

@article{auer2002nonstochastic,
  title={The nonstochastic multiarmed bandit problem},
  author={Auer, Peter and Cesa-Bianchi, Nicolo and Freund, Yoav and Schapire, Robert E},
  journal={SIAM Journal on Computing},
  volume={32},
  number={1},
  pages={48--77},
  year={2002},
  publisher={SIAM}
}

@article{gittins1979bandit,
  title={Bandit processes and dynamic allocation indices},
  author={Gittins, John C},
  journal={Journal of the Royal Statistical Society Series B: Statistical Methodology},
  volume={41},
  number={2},
  pages={148--164},
  year={1979},
  publisher={Oxford University Press}
}

@article{ghavamzadeh2015bayesian,
  title={Bayesian reinforcement learning: A survey},
  author={Ghavamzadeh, Mohammad and Mannor, Shie and Pineau, Joelle and Tamar, Aviv and others},
  journal={Foundations and Trends{\textregistered} in Machine Learning},
  volume={8},
  number={5-6},
  pages={359--483},
  year={2015},
  publisher={Now Publishers, Inc.}
}

@article{robbins1952some,
  title={Some aspects of the sequential design of experiments},
  author={Robbins, Herbert},
  journal={Bulletin of the American Mathematical Society},
  volume={58},
  number={5},
  pages={527--535},
  year={1952}
}

@book{bellman1957dp,
  author    = {Richard Bellman},
  title     = {Dynamic Programming},
  publisher = {Princeton University Press},
  year      = {1957}
}

@article{rothschild1974two,
  title={A two-armed bandit theory of market pricing},
  author={Rothschild, Michael},
  journal={Journal of Economic Theory},
  volume={9},
  number={2},
  pages={185--202},
  year={1974},
  publisher={Elsevier}
}

@book{lattimore2020bandit,
  title={Bandit Algorithms},
  author={Lattimore, Tor and Szepesv{\'a}ri, Csaba},
  year={2020},
  publisher={Cambridge University Press}
}

@article{pearson1909new,
  title={On a new method of determining correlation between a measured character {A}, and a character {B}, of which only the percentage of cases wherein {B} exceeds (or falls short of) a given intensity is recorded for each grade of {A}},
  author={Pearson, Karl},
  journal={Biometrika},
  volume={7},
  number={1/2},
  pages={96--105},
  year={1909},
  publisher={JSTOR}
}

@article{lev1949point,
  title={The point biserial coefficient of correlation},
  author={Lev, Joseph},
  journal={The Annals of Mathematical Statistics},
  volume={20},
  number={1},
  pages={125--126},
  year={1949},
  publisher={Institute of Mathematical Statistics}
}



\end{document}